\documentclass[journal]{IEEEtai}

\usepackage{amssymb}
\usepackage{url}
\usepackage{subfigure}
\usepackage{graphicx}
\usepackage{amsmath}
\usepackage{amsthm}
\usepackage{booktabs}
\usepackage{algorithm}
\usepackage{algorithmic}
\usepackage[switch]{lineno}
\usepackage{multirow}
\usepackage{makecell,balance}

\usepackage{cite}
\usepackage[T1]{fontenc}

\usepackage{xcolor}
\usepackage{tikz}
\newcommand*\emptycircle[1][0.8ex]{\tikz\draw (0,0) circle (#1);}
\newcommand*\halfcircle[1][0.8ex]{%
    \begin{tikzpicture}
    \draw[fill] (0,0)--(90:#1) arc (90:270:#1)--cycle;
    \draw (0,0) circle (#1);
    \end{tikzpicture}
}
\newcommand*\fullcircle[1][0.8ex]{\tikz\fill (0,0) circle (#1);}
\usetikzlibrary{positioning, shapes.geometric}
\usepackage{makecell, tabularx}
\usepackage{ragged2e} 
\newcolumntype{L}{>{\RaggedRight}X} 
\usepackage{lipsum} 
\usepackage{graphicx}

\usepackage{xpatch}

\title{Games for Artificial Intelligence Research:\\A Review and Perspectives}

\author{Chengpeng~Hu,
Yunlong~Zhao,
Ziqi~Wang,
Haocheng~Du,
Jialin~Liu,~\IEEEmembership{Senior Member,~IEEE}
\thanks{The authors are with the Research Institute of Trustworthy Autonomous System, Southern University of Science and Technology (SUSTech), Shenzhen 518055, China. The authors are also with the Guangdong Provincial Key Laboratory of Brain-inspired Intelligent Computation, Department of Computer Science and Engineering of SUSTech.}
\thanks{Y. Zhao is also with LightSpeed Studios, Tencent, Shenzhen 518057, China.}
\thanks{This paper has been accepted by IEEE Transactions on Artificial Intelligence.}
}

\begin{document}

\maketitle

\begin{abstract}
Games have been the perfect test-beds for artificial intelligence research for the characteristics that widely exist in real-world scenarios. Learning and optimisation, decision making in dynamic and uncertain environments, game theory, planning and scheduling, design and education are common research areas shared between games and real-world problems. Numerous open-source games or game-based environments have been implemented for studying artificial intelligence. In addition to single- or multi-player, collaborative or adversarial games, there has also been growing interest in implementing platforms for creative design in recent years. Those platforms provide ideal benchmarks for exploring and comparing artificial intelligence ideas and techniques. This paper reviews the games and game-based platforms for artificial intelligence research, provides guidance on matching particular types of artificial intelligence with suitable games for testing and matching particular needs in games with suitable artificial intelligence techniques, discusses the research trend induced by the evolution of those games and platforms, and gives an outlook.
\end{abstract}

\begin{IEEEImpStatement}
Games have been playing an essential role in the development of AI techniques and education by serving as valuable test-beds. This paper provides a comprehensive review of recent games and game-based platforms for both game playing and design. The platforms discussed in this paper are comprehensively categorised based on factors such as game type, aim, number of agents, observability, programming languages, and the presence of competitions. Furthermore, the paper investigates the challenges posed by games, existing issues with game-based platforms, and the efforts in leveraging natural languages for game research. By offering a detailed roadmap and research insights, this paper serves as a valuable resource for students and researchers, facilitating their exploration of this field. Besides, the paper presents the evolution of games along with the advancement of AI techniques, thereby igniting inspiration for future investigations. Readers from industry will also discover how AI can impact games and how games can impact AI research through this review.
\end{IEEEImpStatement}


\section{Introduction}

\begin{figure*}[t]
    \centering
    \includegraphics[width=1\linewidth]{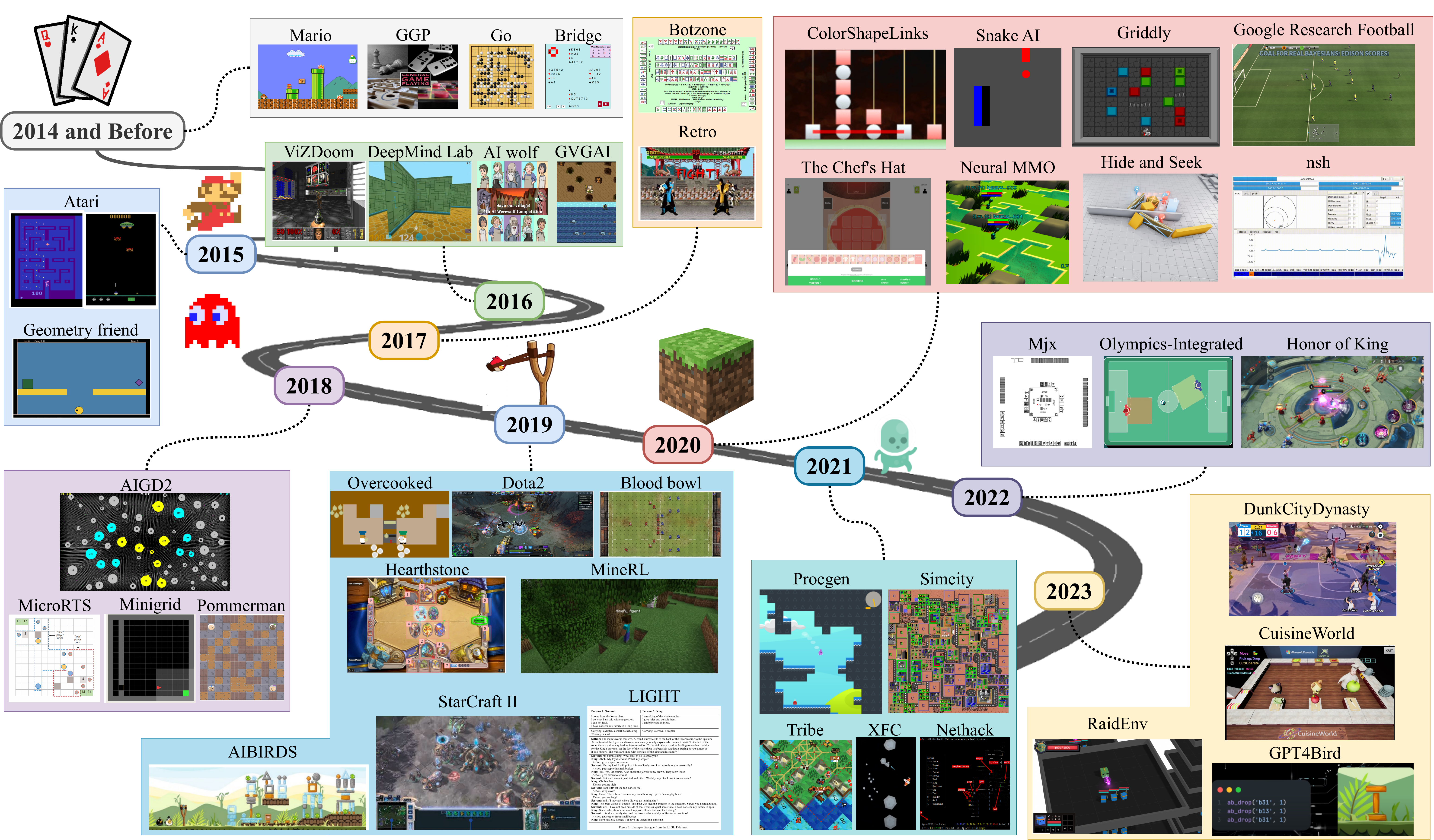}
    \caption{Roadmap of games and platforms, created by authors using screenshots of the games and platforms and figures of papers cited in this work.}
    \label{fig:roadmap}
\end{figure*}

\IEEEPARstart{G}{ames} have played an essential role in the rapid development of artificial intelligence (AI). As a controllable, cost-effective, and realistic simulator~\cite{swiechowski2020game}, games provide virtual worlds for AI to explore. A variety of real-life problems, such as real-time decision making~\cite{mnih2015human,van2016deep}, scheduling~\cite{chinchali2018cellular}, design~\cite{yannakakis2018artificial,liu2021deep} and education~\cite{de2018games}, can be reflected in such virtual worlds. Games have naturally become the perfect test-beds to verify and explore the ability, generality, robustness and safety of AI~\cite{levine2013general,oikarinen2021robust,ji2023safety}. Games or game-based platforms have provided an ideal playground for AI researchers to study, explore, evaluate and experiment with different ideas in a controllable and safe environment~\cite{yannakakis2018artificial,spronck2020artificial}. Such games and platforms also enable us to study and answer what-if questions that are hard to do safely in the real world. They can even stimulate researchers to ask new research questions.

Various games have been used in game-based research platforms, from single-player to multi-player, and from puzzles to real-time strategy (RTS) games, with the emergence of diverse game genres and mechanisms. Such diversity produces a large number of diverse challenges for AI to solve. For instance, it has been demonstrated in various games (e.g., \textit{Chess}~\cite{campbell2002deep}, \textit{Atari}~\cite{mnih2015human}, \textit{Poker}~\cite{bowling2015heads}, \textit{Go}~\cite{silver2017mastering}, 
\textit{StarCraft II}~\cite{vinyals2019grandmaster}, \textit{Dota 2}~\cite{berner2019dota}, \textit{DouDizhu}~\cite{pmlr-v139-zha21a}, Texas hold'em~\cite{zhao2022alphaholdem} and \textit{GuanDan}~\cite{lu2023danzero}) that AI can perform at a level beyond that of humans. Besides playing games by AI, designing new games by AI is an emerging research area which helps exploit AI's potential in creativity~\cite{schnier2021nature}. Games are rich in their content types, including visuals, audio, levels, rules, etc., making them challenging and diverse environments for innovative AI research~\cite{yannakakis2018artificial,risi2020increasing}.

In terms of creative game design, search-based methods, machine learning (ML) and deep learning (DL) techniques have been used for procedural content generation (PCG) where both AI creativity and human-AI co-creativity are explored~\cite{liu2021deep,togelius2011search,shaker2016procedural,Summerville2018Procedural,DeKegel2020Procedural,guzdial2022procedural,yannakakis2023affective}. Emergent abilities of large language models (LLMs) also open up new opportunities and paradigms of playing and designing games~\cite{park2023generative,Kumaran2023scenecraft,Todd2023level,Sudhakaran2023prompt,shyam2024mariogpt,wu2024spring}. More recently, Bruce et al.~\cite{bruce2024genie} from DeepMind introduce~\emph{Genie}, the generative interactive environment that is capable of learning from videos to generate playable, endless 2D platformer games from an image prompt.
AI-generated contents can help to enhance the generality of AI itself~\cite{risi2020increasing}. AI techniques can help to automate the design tasks to accelerate the process, generate novel and diverse contents, and personalise contents for individual players~\cite{liu2021deep,guzdial2022procedural}.

This paper provides a bird's-eye view of publicly available game-based platforms and games for AI research, aiming to provide a brief but comprehensive overview for researchers and students that want to either \textit{use AI for games} or \textit{use games for AI}. Fig. \ref{fig:roadmap} presents the roadmap of games and platforms.
Links to the source code of the reviewed games and platforms, corresponding research papers, and competitions are provided on the GitHub page\footnote{All the links to the reviewed games, platforms and user instructions are collected at \label{alt}\url{https://github.com/SUSTechGameAI/GameAIPlatforms}.}.

The paper is organised as follows. Section \ref{sec:sc} explains the scope and methodology of this survey. Taxonomies of playing and designing tasks involving games for AI research are provided in Sections \ref{sec:gfp} and \ref{sec:design}, respectively. The research trends and challenges induced by the emergence and evolution of those games and game-based platforms are discussed in Section \ref{sec:discuss}. Finally, Section \ref{sec:con} concludes.

\section{Methodology of Survey}
\label{sec:sc}
Our survey considers the interpretation of \textit{games} as Liu
et al.~\cite{liu2021deep}, ``\textit{any games a human would conceivably play, including board games, card games, and any type of video games, such as arcade games, role-playing games, first-person shooters, puzzle games, and many others}''.
To simplify the language, from now on, the term ``\textit{game}'' refers to either a game simulator or an environment with multiple games, while ``\textit{game-based platforms}'' (or ``\textit{platform}'' in short) specially points to the latter. Yannakakis and Togelius~\cite{yannakakis2018artificial} have enumerated and discussed some games for playing and design till 2017. Shao et al.~\cite{shao2019survey} have reviewed some video games for researching deep reinforcement learning (deep RL), which were released before 2019. The work of Swiechowsk~\cite{swiechowski2020game}, Giannakos et al.~\cite{giannakos2020games} and Duan et al.~\cite{duan2022survey} have also touched on this topic. However, to our best knowledge, no existing paper has provided a complete review or taxonomy on recent games.

As a side note, the fields of game AI, autonomous driving,robotics and etc., share some similar challenges~\cite{ibarz2021train,kiran2021deep}. On one hand, some of the games, simulators or virtual environments may be used by one or more fields. On the other hand, the AI approaches and techniques widely used in one field can be applied to another. However, to be more focused in this survey, virtual environments specifically designed for simulating and studying autonomous driving and robotic systems, such as Carla~\cite{dosovitskiy2017carla}, Omniverse~\cite{Nvidia2021Omniverse} and Mujoco~\cite{todorov2012mujoco} with accurate physics simulations, are not covered.

The following survey methodology is applied. First, we have gone through the competitions organised at related conferences in AI or game AI, including NeurIPS/NIPS, IJCAI, AAAI, ICML, ECML, ACML, AIIDE, IEEE CIG, IEEE CoG, ACM FDG, ACG, GECCO, WCCI/CEC, between 2016 and February 2024, and collected corresponding games/ platforms. 
Then, we searched with \emph{Google Scholar} and \emph{Web of Science} using the following search terms, ``game'' AND (``artificial intelligence'' OR ``AI'') AND (``design" OR ``generation'') AND (``platform'' OR ``interface'' OR ``application'' OR ``software''), for related papers published or available online between 2016 and February 2024, assuming that most of the earlier ones can be found in Yannakakis and Togelius~\cite{yannakakis2018artificial}, Shao
et al.~\cite{shao2019survey}, and then investigated if the papers used any game-based platform for AI research. Finally, for all related papers that did not specify the link to the corresponding platform or code, we searched for the code on the authors' personal pages, in \emph{GitHub}, and using \emph{Google Search}. 
The year 2016 is chosen for the following reasons. AlphaGo~\cite{Silver2016MasteringTG} introduced in 2016 is a milestone for AI research. Its success presents the potential of several AI approaches and techniques, such as deep RL and search methods, in mastering complex games with vast search spaces and strategic depth beyond the human level, and inspires not only research on game AI~\cite{meta2022human,wurman2022outracing} but also interdisciplinary research, such as AI for science~\cite{segler2018planning,davies2021advancing} and AI for social interaction~\cite{duenez2023social}. As discussed previously, most of the related games published before 2017 have been covered by~\cite{yannakakis2018artificial,shao2019survey}. However, lots of new games have emerged as shown in this survey.
Again, only publicly available ones are included.
It's worth mentioning that a platform or game may have been available online before the corresponding paper is published. Besides, although some commercial games are not open-source, they provide some APIs to access the games for research, e.g., \textit{StarCraft II}~\cite{vinyals2019grandmaster}. Such games are also included.

\section{Playing Games by AI}
\label{sec:gfp}

\begin{figure*}[t]
    \centering    \includegraphics[width=1\linewidth]{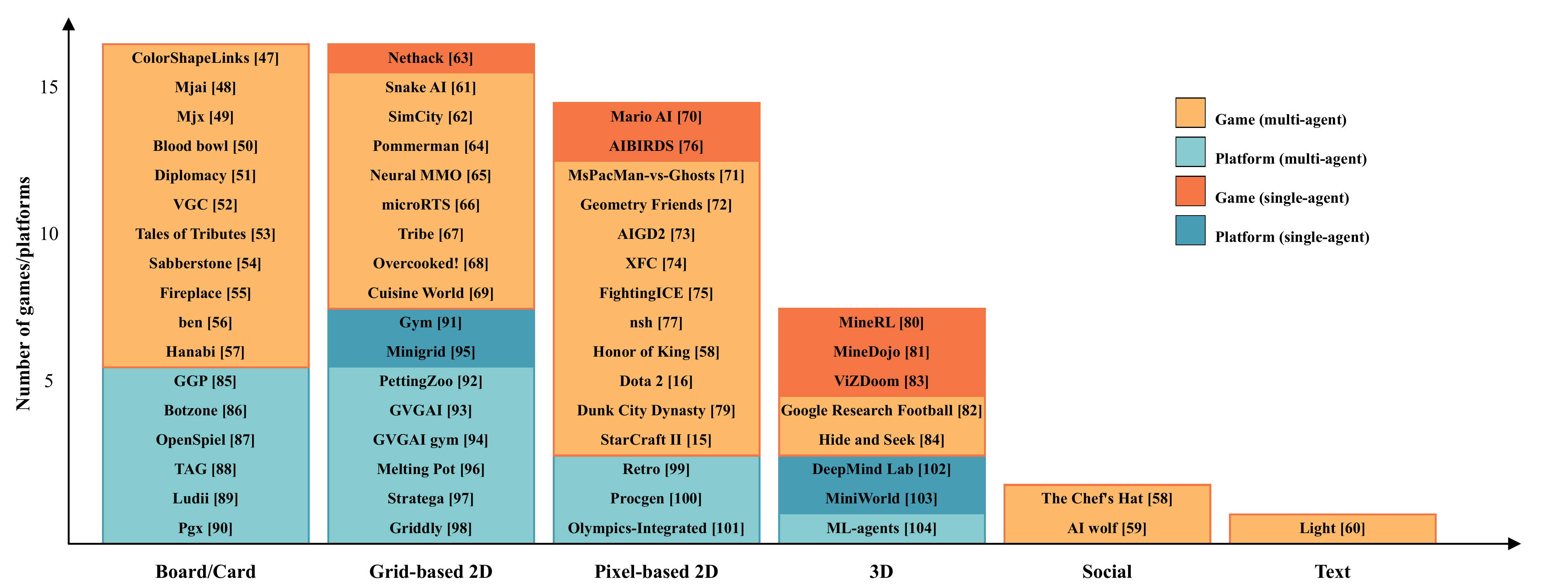}
    \caption{Games and platforms for playing, categorised as board/card, grid-based 2D, pixel-based 2D, 3D, social and text game types.} 
    \label{fig:gameplaying}
\end{figure*}

Games considered in this survey can be briefly categorised into tabletop games and video games. Tabletop games, including puzzles, board games and card games, usually have few requirements for image rendering, use simple objects such as boards, pieces and cards, and do not rely on digital devices. States of such games can be easily represented by structured data. Video games encompass a wide range of genres such as role-playing games, shooter games, racing games, puzzle games, and more. Unlike tabletop games, video games contain visual and audio contents based on digital devices. This poses a challenge of extracting high-dimensional but uninformative states and action features. Moreover, credit assignment in games may be challenging due to the game length, hidden behaviours of players, combinatorial actions and sparse rewards. 
Tab. \ref{tab:playing_single}, Tab. \ref{tab:playing_multiple} and Fig. \ref{fig:gameplaying} categorise the games and platforms according to the game genre, research aims, the number of agents and observability of the simulated physical space with which agent(s) interact, as well as programming language and if they have been used by any AI-related competitions.

\begin{table*}[htbp]
    \centering
    \setlength{\tabcolsep}{4pt}
        \caption{\quad 40 games for playing by AI. The column with header \textbf{C} indicates whether an AI competition is associated.
    \textit{Geometry Friends}, available before 2016, is included as it's used in recent research. 
    ``1'', ``2'' and ``>2'' denote the number of agents, representing single-, two- and multi-agent. \emptycircle, \fullcircle  and \halfcircle denote whether the platform is deterministic, stochastic or both. $^*$ highlights commercial games with publicly available API.}

    \begin{tabular}{c|c|c|cc|ccc|cc|c|c|c}    
        \toprule
   \multirow{2}{*}{} & \multirow{2}{*}{\textbf{Type}}  & \multirow{2}{*}{\textbf{Platform}}  &\multicolumn{2}{c|}{\textbf{Aim}}   & \multicolumn{3}{c|}{\textbf{\# Agents}} & \multicolumn{2}{c|}{\textbf{Observability}}&\multirow{2}{*}{\textbf{Language}}  & \multirow{2}{*}{\textbf{Competition}} &  \multirow{2}{*}{\textbf{Reference}}   \\

   & & & Planning & Learning & $1$ & $2$ & $>2$ & Full & Partial & &\\
    \midrule

    \multirow{13}{*}{{\makecell{\textbf{Tabletop}\\\textbf{Games}\\(Section \ref{sec:classic})}}}
    & \multirow{5}{*}{Board}  & \textit{ColorShapeLinks}    & \checkmark&  &   & A &   & \checkmark&  &C\# & \checkmark    & \cite{fachada2021colorshapelinks} \\
        &   &   \textit{Mjai}   &  \checkmark & &   &  &  A & &  \checkmark&  JavaScript &  &  \cite{mjai}\\
            &   & \textit{Mjx}   &  & \checkmark&   &  & A & & \checkmark& Python &  & \cite{koyamada2022mjx}\\
    &   & \textit{Blood bowl}    &   & \checkmark&  & A &   & \checkmark&   &Python & \checkmark   & \cite{justesen2019blood}   \\

    &   &   \textit{Diplomacy}   &  \checkmark&  \checkmark &   & &   M &  \checkmark&  &   Python &    &  \cite{bakhtin2023mastering} \\
    \cmidrule{2-13}
    & \multirow{5}{*}{Card}  &  \textit{VGC}   &  & \checkmark&   &  A &  & & \checkmark& Python & \checkmark  &     \cite{reis2021vgc}\\
        &   &  \textit{Tales of Tributes}   &  & \checkmark&   & A &  & & \checkmark& C\# & \checkmark  &     \cite{kowalski2023introducing}\\
                &   & \textit{Sabberstone} & \checkmark& \checkmark& & A&   &  & \checkmark& C\# &  \checkmark & \cite{dockhorn2019introducing} \\
                &   &  \textit{Fireplace} &  \checkmark&  \checkmark& &  A&   &  &  \checkmark&  Python &   \checkmark &  \cite{fireplace2014} \\

    &&    \textit{ben}   &  &   \checkmark&   &   &   M& &  \checkmark&   Python &   \checkmark  &       \cite{Dali2022}\\
            &   &  \textit{Hanabi}   &  & \checkmark&   &  A & M & & \checkmark& Python & \checkmark  &     \cite{bard2020hanabi}\\

    \cmidrule{2-13}

    & \multirow{2}{*}{Social}  & \textit{The Chef's Hat}    & & \checkmark & & & A &  & \checkmark & Python & \checkmark     & \cite{barros2020s}\\
    &  &\textit{AI wolf}   &  & \checkmark &   &  & M &  &  \checkmark &Java/Python &  \checkmark  & \cite{toriumi2016ai}\\
    \cmidrule{2-13}
        & \multirow{1}{*} {Text}  &  \textit{Light}    & &  \checkmark & &  M &  M &  \checkmark &  &  Python &     &  \cite{urbanek2019learning}\\

     \midrule

   \multirow{23}{*}{\makecell{\textbf{Video}\\\textbf{Games}\\(Section \ref{sec:video})}} 
        & \multirow{8}{*}{\makecell{Grid-\\based\\2D}} &\textit{Snake AI}    &\checkmark&  & & A&  &  \checkmark&  &Java & \checkmark & \cite{brown2021snakes} \\

       &   &\textit{SimCity}  &  & \checkmark&  \emptycircle&  &    & \checkmark&  & Python &       & \cite{earle2020using}\\
    &   &\textit{Nethack}   & &  \checkmark& \fullcircle &  &   &  & \checkmark&Python &  \checkmark      & \cite{kuttler2020nethack} \\

        & &\textit{Pommerman}   &  & \checkmark&  &  & M & \checkmark&  & Java/Python &   \checkmark & \cite{resnick2018pommerman} \\
    &  & \textit{Neural MMO} &  & \checkmark&  &  & M & \checkmark& \checkmark& Python & \checkmark & \cite{suarez2021neural} \\
    &    & \textit{microRTS}   &\checkmark& \checkmark& & A& &  \checkmark& \checkmark& Java&\checkmark & \cite{ontanon2018first}   \\
    &  &\textit{Tribe}   & \checkmark& \checkmark  & & A& A&  & \checkmark&Java &  & \cite{perez2020tribes} \\
    &  &\textit{Overcooked!}    & \checkmark& \checkmark &  & C&  & \checkmark&  & Python &      & \cite{carroll2019utility}\\
    &  &  \textit{Cuisine World} &  \checkmark  &  \checkmark &  & &     C&  \checkmark &  & C++ &    & \cite{gong2023mindagent}\\ 
    \cmidrule{2-13}

    &  \multirow{12}{*}{\makecell{Pixel-\\based\\2D}} &\textit{Mario AI}    &\checkmark&  & \fullcircle &  &   & \checkmark&  &Java &  \checkmark& \cite{karakovskiy2012mario} \\
            &  & \textit{MsPacMan-vs-Ghosts}    &  \checkmark& & \fullcircle&  &   &   \checkmark&  \checkmark  & Java &  \checkmark &  \cite{williams2016ms} \\
                &  &\textit{Geometry Friends}   & \checkmark&  && C&   & \checkmark& &C\#&  \checkmark & \cite{prada2015geometry} \\
    &  & \textit{AIGD2}& \checkmark& & & A &  &\checkmark & &Java &  &\cite{lucas2018game} \\

        &  &  \textit{XFC} &   &  \checkmark&   \fullcircle &&   &  \checkmark&  & Python &    \checkmark  & \cite{arnett2024x}\\

    &  & \textit{FightingICE} &  & \checkmark& & A&   & \checkmark&  &Python &   \checkmark  
    &\cite{khan2022darefightingice}\\ 

               &  &\textit{AIBIRDS}    & \checkmark& \checkmark&\emptycircle&  &   &  \checkmark&  &Java/Python & \checkmark & \cite{renz2019ai} \\
           &  &   \textit{nsh}  &  &   \checkmark &   \halfcircle   &   &  &  &    \checkmark&    Python/Lua &    &   \cite{nsh-github}\\
    &   & \textit{Honor of King}~$^*$   &  &  \checkmark &   &  A&  &  &  \checkmark&  Python &  \checkmark    &  \cite{wei2022hok_env}\\
    &   & \textit{Dota 2}~$^*$   &  &  \checkmark &  &  &  M &  &  \checkmark& Python/Lua &  \checkmark     &  \cite{berner2019dota}\\
       &  &          \textit{Dunk City Dynasty}~$^*$   &  &   \checkmark &    &   &  M &  &    \checkmark&    Python &    &   \cite{FuXiRL2023DunkCityDynasty}\\
       &  & \textit{StarCraft II}~$^*$   &  &  \checkmark &   &  A &  M&  &  \checkmark& C++/Python &    \checkmark &  \cite{vinyals2019grandmaster}\\

    \cmidrule{2-13}

   &\multirow{5}{*}{3D} &\textit{MineRL}   &  & \checkmark& \halfcircle&  &    & \checkmark& \checkmark& Python &  \checkmark & \cite{guss2019minerldata}\\
      & &\textit{MineDojo}   &  & \checkmark& \halfcircle&  &    & \checkmark& \checkmark& Python &  \checkmark & \cite{fan2022minedojo}\\
            &  &  \textit{Google Research Football} &  &  \checkmark& & &  M  &  \checkmark&  & Python &     &  \cite{kurach2020google}\\ 
    &  &\textit{ViZDoom}   & \checkmark& \checkmark&  \halfcircle&  &    &  & \checkmark& Python &  \checkmark & \cite{kempka2016vizdoom} \\

          &  &  \textit{Hide and Seek} &   \checkmark &  \checkmark&  & &     M& &  \checkmark & Python &    & \cite{baker2019emergent}\\ 

         \bottomrule
 \multicolumn{3}{l}{\emptycircle ~Single-player deterministic game} & \multicolumn{5}{l}{\fullcircle ~Single-player stochastic game} & \multicolumn{3}{l}{\halfcircle Configurable} &\multicolumn{2}{l}{~~~$^*$ Commercial game with API}  \\
\multicolumn{3}{l}{Games with \textbf{\#Agent$\geq2$}:} & \multicolumn{5}{l}{``A'' for adversarial} & \multicolumn{3}{l}{``C'' for cooperative} & \multicolumn{1}{l}{~~~``M'' for mixed} 
    \end{tabular}

    \label{tab:playing_single}
\end{table*}

\begin{table*}[htbp]
    \centering\setlength{\tabcolsep}{4pt}
        \caption{20 platforms with multiple games for playing by AI (cf. caption of Tab. \ref{tab:playing_single} for reading instructions). Whether a game is adversarial or collaborative is not indicated as each platform contains a number of different games. \textit{GGP}, available before 2016, is included as it's widely used in recent research. 
    ``1'', ``2'' and ``>2'' denote the number of agents, representing single-, two- and multi-agent.
    Action-adventure game (AAG) involves video games such as puzzles and combat games. 
    } 
    \begin{tabular}{c|c|c|ccccc|ccc|cc|c|c|c}    
        \toprule
        \multirow{2}{*}{\textbf{Platform}}  &\multicolumn{2}{c|}{\textbf{Aim}}  & \multicolumn{5}{c|}{\textbf{Game genre}} & \multicolumn{3}{c|}{\textbf{\#Agents}}  & \multicolumn{2}{c|}{\textbf{Observability}}&\multirow{2}{*}{\textbf{Language}}  & \multirow{2}{*}{\textbf{Competition}} &  \multirow{2}{*}{\textbf{Reference}}   \\

    & Planning & Learning & Board& Card& AAG &RTS& Sport&$1$ & $2$ & $>2$ & Full & Partial & & &\\
       \midrule
\textit{GGP}     & \checkmark &  & \checkmark & \checkmark  & & & &  \checkmark &\checkmark &\checkmark & \checkmark  & & Java & \checkmark & \cite{genesereth2005general}\\

\textit{Botzone}      & \checkmark& \checkmark&  \checkmark & \checkmark &  & & & & \checkmark& \checkmark& \checkmark& \checkmark&  Java/Python &  \checkmark& \cite{zhou2017botzone} \\
\textit{OpenSpiel}  & \checkmark& \checkmark& \checkmark & \checkmark &  & & & \checkmark& \checkmark& \checkmark & \checkmark& \checkmark& C++/Python &   & \cite{lanctot2019openspiel}    \\
\textit{TAG} & \checkmark & & \checkmark & \checkmark &  & & &   & \checkmark& \checkmark& \checkmark& \checkmark&Java &   & \cite{gaina2020tag}  \\
\textit{Ludii}  & \checkmark& \checkmark& \checkmark & & & & & &  \checkmark &  & \checkmark  & & Java/Python & \checkmark&\cite{stephenson2019ludii}\\

 \textit{Pgx} &  \checkmark &  \checkmark &   \checkmark &   \checkmark &  & & &   \checkmark  &   \checkmark&   \checkmark&   \checkmark&   \checkmark& Python &   &  \cite{koyamada2024pgx}  \\

\midrule
   \textit{Gym}  & \checkmark& \checkmark&  & &\checkmark & & &\checkmark&  &   &\checkmark& \checkmark& Python &     & \cite{brockman2016openai}\\
    \textit{PettingZoo}   & & \checkmark& \checkmark & \checkmark& \checkmark&  & & \checkmark & \checkmark& \checkmark& \checkmark& \checkmark&Python &     & \cite{terry2021pettingzoo}\\

\midrule

\textit{GVGAI}     & \checkmark& & & &\checkmark&&&\checkmark& \checkmark& \checkmark & \checkmark& & Java/Python & \checkmark & \cite{perez2016general} \\
\textit{GVGAI gym}     & \checkmark& \checkmark& &&\checkmark&&&\checkmark& \checkmark& \checkmark & \checkmark& & Java/Python & \checkmark & \cite{perez2019general} \\
\textit{Minigrid}    & & \checkmark & & & \checkmark& & &  \checkmark&  &   &  \checkmark& \checkmark&Java/Python &     & 
    \cite{minigrid}\\
\textit{Melting Pot}  &  & \checkmark&  & & \checkmark& & & &  & \checkmark& \checkmark& \checkmark& Python/Lua&    & \cite{leibo2021meltingpot}\\
\textit{Stratega}    &\checkmark& \checkmark & &&&\checkmark&& & \checkmark& \checkmark&  & \checkmark& C++/Python&    & \cite{dockhorn2020stratega} \\
\textit{Griddly}   & \checkmark& \checkmark & &&\checkmark&\checkmark&&\checkmark& \checkmark& \checkmark & \checkmark& \checkmark&Python &    & \cite{bamford2021griddly} \\

\midrule

\textit{Retro}  &  & \checkmark& &&\checkmark&&&\checkmark  & & & \checkmark& \checkmark&Python &  \checkmark & \cite{bhonker2017playing} \\
\textit{Procgen}    &  & \checkmark& &&\checkmark&&&\checkmark&  &   & \checkmark&  & Python & \checkmark  & \cite{cobbe2020leveraging}\\
\textit{Olympics-Integrated}   &  & \checkmark& &&&&\checkmark& & \checkmark&   & \checkmark&  &Python &   \checkmark  & \cite{Zhang2022olympics}\\
\midrule

\textit{DeepMind Lab}    &  & \checkmark& &&\checkmark&&& \checkmark&  &   & \checkmark& \checkmark& Python &    & \cite{beattie2016deepmind}\\

\textit{MiniWorld}   &  & \checkmark& &&\checkmark&&&\checkmark&  &   & \checkmark& \checkmark& Python &    & 
   \cite{gym_miniworld}\\
   \textit{ML-agents}   &  & \checkmark& &&\checkmark&&\checkmark& \checkmark& \checkmark& \checkmark & \checkmark& \checkmark& Python &   & \cite{juliani2018unity}\\

         \bottomrule
    \end{tabular}
    \label{tab:playing_multiple}
\end{table*}
\subsection{Tabletop Games}
\label{sec:classic}

Board, card, puzzle games and other tabletop games raise several challenges to AI, such as partial observability and multi-agent cooperation or competition.

In \textit{ColorShapeLinks}~\cite{fachada2021colorshapelinks}, players win by connecting pieces with the same colour or shape. The platform of \textit{ColorShapeLinks} provide not only a visual front-end via the \textit{Unity} game engine, but also a pure text-based one.
\textit{ColorShapeLinks} has been used for education and a testbed of an AI competition. It can be easily accessed by individuals with different backgrounds.
However, \textit{ColorShapeLinks} is relatively simple for research purposes since good solutions can be quickly found.
\textit{Mjai}~\cite{mjai} and \textit{Mjx}~\cite{koyamada2022mjx} are partial-observable AI environments of \textit{riichi Mahjong}. Thirteen tiles on the own hand and discarded cards are available for the player. If a special permutation of cards is formed, the player wins the game. \textit{Mjx} is 100 times faster than \textit{Mjai} in running time and is compatible with \textit{Gym}'s API~\cite{koyamada2022mjx}.
\textit{Blood bowl}~\cite{justesen2019blood} simulates a fully-observable board game that two players control their own teams and are rewarded with touchdowns. At each turn, the player moves their avatars to block the team of the opposing player or pass the ball. The environment is compatible with \textit{Gym}'s API for RL research~\cite{justesen2019blood}.
In \textit{Blood bowl}, the observation space consists of both spatial and non-spatial information and the action space differs across states, which leads to a new challenge in representation.
\textit{Diplomacy} is a seven-person board game, considering both cooperation and competition~\cite{bakhtin2023mastering}. In this game, AI agents are required to understand human players' imitations and behaviours.

\textit{Pokémon Video Game Championships (VGC)} is a turn-based strategy game~\cite{reis2021vgc}. Players need to collect Pokémons with different abilities to build a team. During the battles, player agents decide the order of Pokémons and specify the skills to attack the opponents according to previous information.
The work of~\cite{kowalski2023introducing} introduces a deck-building card game, \textit{Tales of Tribute}. Players build the deck with the same initial cards during the game through battles and trades, meeting some dynamics in a long-term strategy. 
\textit{Tales of Tribute} also provides a graphic interface and supports implementing AI agents with \texttt{C\#}.
\textit{Hearthstone: Heroes Of Warcraft} is a popular commercial turn-based CCG. There are open-source implementations, e.g., \textit{Sabberstone}~\cite{dockhorn2019introducing} and \textit{Firepalce}~\cite{fireplace2014}, for playing.
The three games are all collectable card games (CCG), where players build decks and fight against opponents, but they have different research purposes. 
\textit{Sabberstone}, \textit{Firepalce} and \textit{Tales of Tribute} focus on researching decision-making agents, i.e., how to play cards. However, \textit{Tales of Tribute} can also be used to study building decks by AI. Meanwhile, \textit{VGC} involves meta-game balance that adjusts attributes of cards during the battles.
\textit{Bridge} is another challenging board game for AI because of its partial observability, high branching factor, cooperation and competition~\cite{ventos2017bridge}. 23 editions of the World Computer-Bridge Championship have been organised with the latest one in 2019\footnote{World Computer-Bridge Championship: \url{https://bridgebotchampionship.com/}}. The game engine of the championship was not found by authors. However, there is a recently released game engine called \textit{ben}~\cite{Dali2022} for AI research in \textit{Bridge}.
Different from some traditional card games, players of \textit{Hanabi}~\cite{bard2020hanabi} can only see the cards of other players. During the gameplay, the player can tell others some hints about their cards. As a fully cooperative game, players need to form fireworks that consist of the same colour cards in ascending order to win.

In most games, agents directly control avatars to move or play cards, while in some others, agents take non-direct actions such as languages.
The social card game, \textit{The Chef's Hat}~\cite{barros2020s} is also implemented based on \textit{Gym}~\cite{brockman2016openai} and used for RL research. The winning goal of the game is to discard all hand cards~\cite{barros2020s}.
\textit{AI wolf}~\cite{toriumi2016ai} is a multi-agent communication game that aims to find the wolf by talking with players. The coordination of multiple agents in partial observation only by talking is considered in the platform.
\textit{Light} is a text-based adventure game~\cite{urbanek2019learning}, with which dialogue-based human-AI interaction is studied.

Usually, no specific design is required for representing states in tabletop games, while the action space in those games changes during an episode. For example, in the game of \textit{Go}, the game state is easily constructed by a matrix, where three symbols or digits denote white, black or empty. However, the empty space for placing stones changes in a decreasing trend, leading to a variable action space. On the other hand, tabletop games often involve multiple players with complex and even dynamic relationship (collaboration vs. competition), pose new challenges and encourage studies in various research fields such as multi-agent RL~\cite{littman1994markov,bucsoniu2010multi,zhang2021multi}, game theory~\cite{fudenberg1991game,hazra2022applications} and psychology~\cite{boyle2011role}.

\subsection{Video Games}
\label{sec:video}
This section categorises video games into grid-based 2D, pixel-based 2D and 3D according to the simulated physical space with which avatar(s) interact. 
Grid-based games are usually constructed by some preset tiles where players act with a set of discrete actions. Unlike grid-based games, pixel-based games feature continuous state space with either continuous or discrete action space. The difference in the state representation results in different feature extraction between grid-based and pixel-based games. The state in grid-based games can be easily converted to a low-dimensional matrix since the number of tiles is usually countable. 
On the other hand, feature extraction for pixel-based games is usually applied to states represented by high-dimensional images.
3D games extend 2D games with an additional height dimension, which not only enriches the visual and interactive depth of games but also introduces significant challenges in representing the state effectively. Besides, the action spaces of 3D games are usually more complicated or larger than those of 2D games.

The transition from grid-based 2D to pixel-based 2D and 3D games marks a significant evolution in video games during the development of technology. This progress poses new challenges and opportunities for AI research. As games become more sophisticated and closer to the real-world, AI techniques are evolving to meet these challenges simultaneously. Furthermore, the difference between games and 3D virtual environments for robotics and autonomous driving, such as \textit{MuJoCo}~\cite{todorov2012mujoco} and \textit{CARLA}~\cite{dosovitskiy2017carla}, is narrowing.

\subsubsection{Grid-based 2D Games}
Classic single-player and multi-player competitive games have been implemented using grid-based maps.
\textit{Snake AI}~\cite{brown2021snakes} allows two snakes to fight for an apple. Since the apple is placed randomly in maps, the game can sometimes be unbalanced.
\textit{SimCity}~\cite{earle2020using} is an open-ended city-building 2D video game with an open-source version provided for RL research.
\textit{Nethack}~\cite{kuttler2020nethack} is a single-agent roguelike game with procedurally generated levels. Agents meet the different scenarios at each episode, which drives AI's abilities for long-term exploration and generalisation. Compared with existing single-player games, \textit{Nethack} provides a faster simulator with massive tasks and game dynamics.
Multi-agent games involve multiple independent decision agents, oriented towards cooperative and competitive behaviours.
\textit{Pommerman}~\cite{resnick2018pommerman} is a multi-player competitive game for at least four players, who start with a limited number of bombs. Players move in a grid-based map and plant the bombs to destroy their opponents. 
\textit{Pommerman} allows multiple players to cooperate and compete as a team on the battlefield.
\textit{Neural Massive Multiplayer Online (Neural MMO)}~\cite{suarez2021neural} considers a large multi-agent scenario where at most 1024 agents learn to cooperate and compete to survive in the grid-based world at the same time.

Modern strategy games incorporate digital elements to achieve complex and diverse mechanisms. 
\textit{microRTS}~\cite{ontanon2018first} is a lightweight, grid-based RTS game framework that has fewer rendering sprites and focuses on the game's mechanics, in which units are represented by different colours and shapes. Compared with \textit{StarCraft II}, \textit{microRTS} is easier to manipulate, so that researchers can focus more on the design of AI ideas and algorithms.
As an implementation of \textit{The Battle of Polytopi}, \textit{Tribe}~\cite{perez2020tribes} involves multiple multi-agent scenarios as a turn-based strategy game framework, in which a forward model is supported for decision. Partial observability mode is supported by the framework, where the unexplored area is covered by the fog of war.

The work of Carroll et al.~\cite{carroll2019utility} provides a simple implementation of the commercial game \textit{Overcooked!} for human-AI coordination, in which agents and players work together to cook meals. Agents and players need to move discretely and avoid colliding with each other, which introduces a new challenge of making motion coordination strategies. However, this implementation~\cite{carroll2019utility} encodes the game map as a 2D array, which can hardly express the complete information of \textit{Overcooked!}. For instance, the representation of the game level can not distinguish whether an onion is on a table or the floor. Besides, the ingredients are fixed in the implementation.
It's worth mentioning that the work of \cite{gong2023mindagent} also introduces a game engine similar to \textit{Overcooked!}, called \textit{CuisineWorld}, to evaluate multi-agent cooperation performance. The platform has been tested by LLM agents~\cite{gong2023mindagent}. Though \textit{CuisineWorld} is a 3D game engine, to our best knowledge, 2D game observations are used in \cite{gong2023mindagent}. Instead of traditional representation based on visual information, \textit{CuisineWorld} inverts the game state with text-based representation and the action space consists of several instructions like moving and putting. Compared to the implementation by~\cite{carroll2019utility}, \textit{CuisineWorld} is more suitable for research on LLMs.

Grid-based games usually have simplified state and action space, as well as clear game mechanisms, which allow to train and test AI agents at a faster speed. However, this simplicity limits its usage in AI approaches considering continuous decision space and complex strategies.

\subsubsection{Pixel-based 2D Games} 
Pixel-based games support more complex game mechanisms with continuous state space, while new challenges are introduced.
\textit{Mario AI} framework~\cite{karakovskiy2012mario} is based on the game \textit{Super Mario Bros.}, where players control Mario to run and jump through some barriers in a partially observable environment to arrive at the endpoint. However, the provided levels are simple and a limited number of levels are available, planning algorithms such as A$^*$ can quickly solve it without leveraging sophisticated AI approaches.
\textit{MsPacMan-vs-Ghosts}~\cite{williams2016ms} is the implementation of \textit{Ms. Pac-Man} arcade game, in which the player controls \textit{Ms. Pac-Man} to collect pills while avoiding being hunted by four ghosts. Both full and partial observability are supported in \textit{MsPacMan-vs-Ghosts}.
\textit{Geometry Friends}~\cite{prada2015geometry} provides a cooperative scenario that two agents aim to solve puzzles together, restricted by real physics conditions such as friction and gravity.
A simple implementation of \textit{Planet Wars}, named \textit{AIGD2}~\cite{lucas2018game}, brings a new challenge with extra-long skill-depth to RTS games. 
\textit{XFC}~\cite{arnett2024x} simulates the game \textit{Asteroid Smashers}, in which AI controls a spacecraft to avoid asteroids and destroy pesky rocks. \textit{XFC} designs a specific agent architecture, which is suitable for research on explainable AI and fuzzy AI.
\textit{FightingICE}~\cite{khan2022darefightingice} considers a 2D fighting video game that AI agents fight against opponents according to sound. ``Blind" agents face challenges such as partial observability and insufficient training.
\textit{AIBIRDS}~\cite{renz2019ai} is the platform with the popular puzzle game \textit{Angry Birds}. Agents are required to shoot pigs to destroy the building made of blocks of different types and shapes.

\begin{figure}[t]
    \centering
    \includegraphics[width=0.9\linewidth]{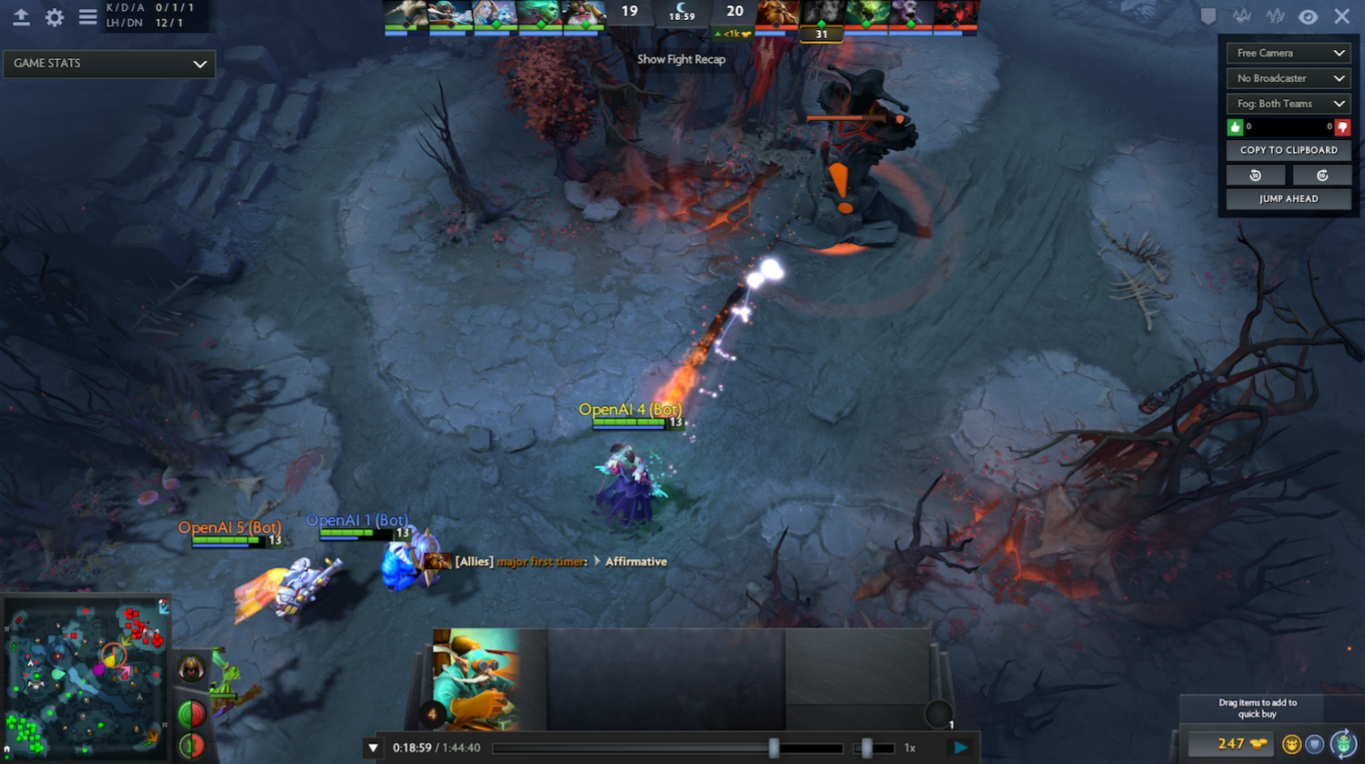}
    \caption{Screenshot of OpenAI Five~\cite{berner2019dota}.}
    \label{fig:dota}
\end{figure}

\textit{nsh}~\cite{nsh-github} is a fight game simulator for the commercial massive multiplayer online (MMO) game \textit{Justice Online}. Renders and memory usage are largely reduced in \textit{nsh}, while game rules are close to the ones in the original game.
\textit{Honor of Kings Arena}~\cite{wei2022hok_env} provides an interface to interact with \textit{Honor of Kings 1v1}, in which agents learn to fight against diverse opponents. An API of \textit{Dota 2} written with \texttt{Lua} language is available for investigating multi-agent RL~\cite{berner2019dota}. Fig.~\ref{fig:dota} presents the heroes controlled by OpenAI Five~\cite{berner2019dota}.
\textit{Dunk City Dynasty} simulates a 3v3 basketball commercial game with the same name~\cite{FuXiRL2023DunkCityDynasty}. A dataset with 400GB human data and records is also provided.
An interface connecting to \textit{StarCraft II}~\cite{vinyals2019grandmaster} is also released for research in multi-agent RL and map design.
Source codes of commercial games are often inaccessible, which means AI researchers can hardly edit commercial games to meet some particular requirements for research, such as creating an abstract, faster forward model of the game, even though their APIs are sometimes available.

\subsubsection{3D Video Games}
3D video games provide a more realistic simulation of the real world. Decision agents need to take actions based on the 3D vision instead of a flat plane in 2D games, which requires effectively processing high-dimensional visual information to make effective decisions.

\textit{MineRL}~\cite{guss2019minerldata} is based on \textit{Minecraft}, where a large-scale human dataset is provided. 
\textit{MineDojo}~\cite{fan2022minedojo} provides a \textit{Gym}-like environment based on \textit{Minecraft}. Agents are supported to directly access the game for multiple tasks including material mining, crafting and building.
Besides, a massive multimodal dataset, captured from video websites, gameplays, wiki pages and Reddit posts is provided.
\textit{Google Research Football}~\cite{kurach2020google} provides a real-world football challenge with a physical 3D engine, in which multiple tasks with variable difficulty are available.
\textit{ViZDoom}~\cite{kempka2016vizdoom} is a highly customisable doom-based environment with partial observability. Raw visual screens from the first-person perspective are received by the agents. 
\textit{ViZDoom} is lightweight and fast. Its integrated editor allows users to manually create diverse task suits. \textit{ViZDoom} is suitable as the test-bed for visual RL research.
The work of Baker et al.~\cite{baker2019emergent} presents a specific platform for \textit{Hide and Seek}, which can be used for training and understanding multi-agent cooperative and competitive AI.

\begin{figure*}[htbp]
    \centering
\includegraphics[width=1\linewidth]{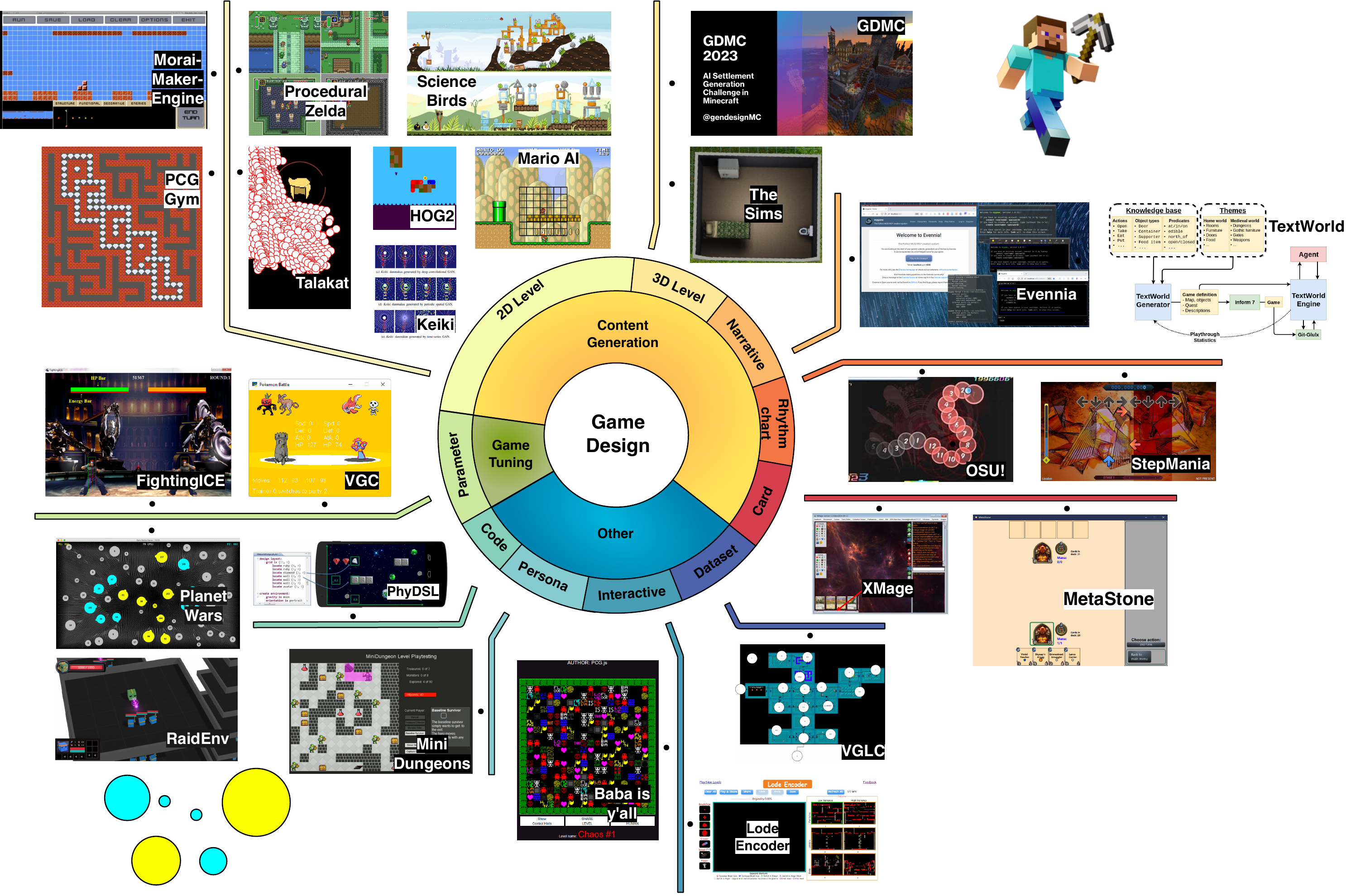}
    \caption{Summary of games and platforms used for studing AI for game design, created by authors using screenshots of the games and platforms, as well as figures from their corresponding papers, reviewed in Section \ref{sec:design}. }
    \label{fig:gamedesign}
\end{figure*}

\subsection{Platforms}\label{sec:platforms}

Classic and modern board, card and puzzle games are often integrated into a single platform as the AI test-bed. 
For example, \textit{General Game Playing} (\textit{GGP})~\cite{genesereth2005general} is designed to evaluate AI's ability in playing different kinds of tabletop games, where rules and states of games are defined with game description language, and has been widely used in AI research and education. Hundreds of single- and multi-player games, such as \textit{Go}, \textit{Tic-Tac-Toe}, \textit{reversi} and \textit{minesweeper}, are available in \textit{GGP}.
\textit{Botzone}~\cite{zhou2017botzone} focuses more on large multi-player games and is famous for its Mahjong AI Competition. \textit{Botzone} also support 2- and 6-person Texas hold'em.
 Besides some popular classic games, \textit{OpenSpiel}~\cite{lanctot2019openspiel} provides \textit{Poker} and \textit{Bridge}, as well as some classic algorithms and visual tools.
Different from \textit{GGP} and \textit{Botzone}, \textit{Tabletop Games Framework} (TAG)~\cite{gaina2020tag} is designed to study general AI in modern tabletop games and has implemented more than 20 games, including \textit{Uno}, \textit{Love Letter} and \textit{Virus!}. An extra analysis tool is equipped in \textit{TAG} for studying action space, branching factors, hidden information, and so on.
 \textit{Ludii}~\cite{stephenson2019overview} is another general game-playing system based on similar principles to \textit{Ludi}~\cite{browne2010evolutionary} but with different mechanisms. It includes various types of games such as board games, boardless games, stacking games, single- and multi-player games. \textit{Ludii} uses \textit{ludemes} as game description grammar. The description language is easy to read and understand according to~\cite{stephenson2019overview}.
\textit{Pgx}~\cite{koyamada2024pgx} is an environment of board games for RL research. It is written in \texttt{JAX} and can do parallel simulations via GPU/TPU. \textit{Pgx} can hold single-/multi-agent board games and Atari-like environments including \textit{Shogi}, \textit{Go}, \textit{Chess}, and \textit{Backgammon}. According to \cite{koyamada2024pgx}, \textit{Pgx} can perform 10-100 times higher throughput compared to other \texttt{Python} libraries.

\textit{OpenAI Gym}~\cite{brockman2016openai} and \textit{PettingZoo}~\cite{terry2021pettingzoo} offer a number of grid- and pixel-based 2D games. \textit{OpenAI Gym}~\cite{brockman2016openai} provides an API standard for RL research, and contains massive games (e.g., \textit{Atari}), and a wide range of third-party environments. However, \textit{Gym} typically focuses on the single-agent setting. To conduct research in multi-agent RL, \textit{PettingZoo}~\cite{terry2021pettingzoo} extends the API standard of \textit{Gym} and integrates some famous environments like multi-player \textit{Atari} games, classic card/board games and multi-agent particle environment~\cite{mordatch2017emergence}. As providing the standard API, \textit{Gym} and \textit{PettingZoo} are now the most used and compatible platforms, which enhance the evaluation and reproduction in the RL community with a higher speed.

\textit{General Video Games AI (GVGAI)}~\cite{perez2016general} platform and \textit{GVGAI gym}~\cite{perez2019general} offer a larger number of grid-based single- and two-player games, and have been used in their associated single-player planing, two-player planing, and single-player learning competitions. Users can create new games and levels in \textit{GVGAI} in an intuitive way using \textit{Video Game Description
Language (VGDL)}~\cite{schaul2013pyvgdl}.
\textit{Minigrid}~\cite{minigrid} is a single-agent platform compatible with \textit{OpenAI Gym}. The included games are simple in rendering content but can implement complex mechanisms.
\textit{Melting Pot}~\cite{leibo2021meltingpot} is a platform with more than 50 diverse games for evaluating the generalisation of multi-agent RL. 
\textit{Stratega}~\cite{dockhorn2020stratega} evaluates the generalisation of strategy agents with multiple games and levels with an $n$-player turn-based setting. Profiling and logging tools are integrated for analysis. Usually, grid-based platforms are highly scalable in terms of game mechanisms and levels, so researchers can easily design specific tasks to address their research problems with the help of grid-based game platforms.
\textit{Griddly}~\cite{bamford2021griddly} particularly provides grid-based real-time strategy games with various numbers of players, with a faster running speed and lower memory than \textit{Minigrid} and \textit{GVGAI}.

The \textit{Retro Learning Environment}~\cite{bhonker2017playing} extends \textit{Gym}, by incorporating popular video games from Super Nintendo Entertainment and Sega Genesis.
\textit{Procgen}~\cite{cobbe2020leveraging} evaluates the efficiency and generalisation of RL with 16 procedurally generated \textit{Gym} environments, the initial state of which can vary for each episode.
\textit{Olympics-Integrated}~\cite{Zhang2022olympics} poses the challenge of designing one single agent to participate in various two-player Olympics games. Since the integrated games in \textit{Retro}, \textit{Procgen} and \textit{Olympics-Integrated} are pixel-based, challenges like feature extraction are introduced for AI agents. Besides, these pixel-based game platforms usually have lower scalability in adding new games and tasks than grid-based platforms such as \textit{GVGAI}.

\textit{DeepMind Lab}~\cite{beattie2016deepmind} acts as a 3D learning environment which is based on some open-source software such as \textit{ioquake3}. It provides a set of challenging 3D navigation and puzzle-solving tasks for agents to learn. The primary purpose of \textit{DeepMind Lab} is to act as a working platform for results in AI, especially deep RL.
\textit{MiniWorld}~\cite{gym_miniworld} provides a more accessible way to construct 3D environments for RL and robotics research.
It is qualified to generate and simulate environments with rooms, doors, mazes and other objects.
\textit{MiniWorld} is simpler than \textit{DeepMind Lab} and \textit{ViZDoom} in the view of engineering, so beginners can get started quickly without mastering Unity3D.
\textit{ML-agents}~\cite{juliani2018unity} is an open-source platform that allows to create new games in a simpler and more flexible manner with the help of the Unity3D toolkit.

\subsection{Guidance for Playing Games}
According to the aim, number of agents, observability, language and competition, games and platforms reviewed in this work are categorised in Tabs.~\ref{tab:playing_single} and~\ref{tab:playing_multiple}. Those games and platforms are often developed for specific research purposes. 
Readers can choose suitable games and platforms according to the categories. For example, if one focuses on learning-based research like RL~\cite{sutton2018reinforcement} and imitation learning~\cite{hussein2017imitation}, games and platforms with \checkmark under \textbf{Learning} in the column \textbf{Aim} of Tabs.~\ref{tab:playing_single} and~\ref{tab:playing_multiple} may be good choices. Examples of related research using those platforms include the work of Mnih \textit{el al.}~\cite{mnih2015human} using \textit{Atari}, and Vinyals \textit{el al.}~\cite{vinyals2019grandmaster} using \textit{StarCraft II}. For research on searching algorithms such as Monte-Carlo tree search~\cite{browne2012survey,swiechowski2023monte}, games and platforms marked in \textbf{Planning} column are good choices.
With respect to different scenarios considering the number of agents, column \textbf{\#Agents} provides such information. Single-agent scenarios usually meet deterministic, stochastic and configurable environments, denoted with \emptycircle, \fullcircle~ and \halfcircle, respectively.
Sparse reward~\cite{bellemare2016unifying,pathak2017curiosity} and long-term exploration~\cite{osband2016deep} are common research topics involved with these scenarios.
Besides, three multi-agent cases including adversarial, cooperative and mixed scenarios are denoted by ``A", ``C" and ``M", respectively. 
Research on game theory~\cite{fudenberg1991game,hazra2022applications} and multi-agent RL~\cite{littman1994markov,bucsoniu2010multi,zhang2021multi} can benefit from the information for choosing suitable testbeds for experiments.
\textbf{Observability} of games and platforms can also be checked in Tabs.~\ref{tab:playing_single} and~\ref{tab:playing_multiple}, which helps with research in partially observable planning~\cite{bertoli2006strong} and RL~\cite{kurniawati2022partially}.

Card/board-related games and platforms, including \textit{Blood bowl}~\cite{justesen2019blood}, \textit{Sabberstone}~\cite{dockhorn2019introducing} and \textit{Botzone}~\cite{zhou2017botzone}, are suitable for research on search, causality and competitive scenarios in multi-agent systems.
\textit{GVGAI}~\cite{liebana2020general}, \textit{Stratega}~\cite{dockhorn2020stratega} and \textit{Procgen}~\cite{cobbe2020leveraging} are developed for the purpose of evaluating the generalisation of AI agents. Both \textit{GVGAI} and \textit{Stratega} can be used for research on planning and learning, while \textit{Procgen} specially focuses on RL.

In the single-agent setting, \textit{Nethack}~\cite{kuttler2020nethack} considers how an AI agent solves exploration, planning, and inventory management problems in a procedurally generated 2D environment. Both \textit{ViZDoom}~\cite{kempka2016vizdoom} and \textit{MineRL}~\cite{guss2019minerldata} require AI agents to explore the game with 3D observation and action space and support RL. \textit{ViZDoom} has some clear goals and limited accessible area, while \textit{MineRL} poses challenges that agents need to explore an open world for survival.
For the multi-agent setting, \textit{Snake AI}~\cite{brown2021snakes}, \textit{microRTS}~\cite{ontanon2018first}, \textit{AIGD2}~\cite{lucas2018game} and \textit{Tribe}~\cite{perez2020tribes} provide test-beds for competitive scenarios. However, \textit{Snake AI} and \textit{AIGD2} are limited to planning. 
Both \cite{carroll2019utility} and \cite{gong2023mindagent} implement \textit{Overcooked!} to study the cooperative behaviours of multiple AI agents. Nevertheless, the former considers how agents cooperate via multi-agent RL, while the latter applies the text-based representation in the game and aims to plan the high-level actions of agents such as moving by LLMs. Moreover, \textit{Neural MMO}~\cite{suarez2021neural}, \textit{Dota2}~\cite{berner2019dota} and \textit{Hide and Seek}~\cite{baker2019emergent} are suitable for learning-based AI in the mixed multi-agent setting with partial observability. \textit{Neural MMO} also consider the large-scale case in terms of the number of agents, compared with the other two games.

Besides, the integrated tools differ between games and platforms. For example. in \textit{GVGAI}, users can easily construct their games and levels via VGDL. \textit{Pommerman}~\cite{resnick2018pommerman} supports highly self-defined maps. However, game levels in \textit{Procgen} can only be generated randomly.
It is recommended to get started from \textit{Gym}~\cite{brockman2016openai} and \textit{PettingZoo}~\cite{terry2021pettingzoo} for single- and multi-agent RL research since they are the most commonly used platforms.  
Some platforms such as \textit{DeepMind Lab}~\cite{beattie2016deepmind} and \textit{ML-agent}~\cite{juliani2018unity} usually are equipped with some visual tools and loggers to analyse the behaviour of agents, but the requirement to learn about the game engines like \textit{Unity} is crucial.

From the two tables, \texttt{Python} is the most used programming language, which aligns with the trend of the famous \texttt{Python} libraries such as \texttt{Torch} and \texttt{TensorFlow}.

\section{Designing Games by AI\label{sec:design}}

\begin{table*}[!ht]
    \centering\setlength{\tabcolsep}{10pt}
        \caption{26 environments for designing games by AI. Although \textit{Mario-AI}, \textit{HOG2}, \textit{PhyDSL}, \textit{StepMania}, \textit{OSU!} and \textit{MiniDungeons} are available before 2016, they have been used for testing and evaluating the procedurally generated \textit{Super Mario Bro} levels in recent years.}
    \begin{tabular}{c|c|c|c|c|c}
        \toprule
    & \textbf{Type} & \textbf{Environment} & \textbf{Language}  & \textbf{Competition}& \textbf{Reference}    \\
    \midrule
        \multirow{4}{*}{{\makecell{\textbf{Game Tuning}\\(Section \ref{sec:tuning})}}}&\multirow{4}{*}{Parameter}  & \textit{FightingICE} &Java  & \checkmark & \cite{lu2013fighting}\\
 &  &  \textit{AIGD2}  &Java& & \cite{lucas2018game} \\

     &  &  \textit{VGC}  &Python& \checkmark & \cite{reis2021vgc} \\
          &  &   \textit{RaidEnv}  & Python&  & \cite{jeon2023raidenv} \\
    \midrule

        & \multirow{9}{*}{2D level} & \textit{Science Birds}  &Java  &  & \cite{ferreira2014a} \\
    \multirow{18}{*}{{\makecell{\textbf{Content Generation}\\(Section \ref{sec:content})}}} &  & \textit{Mario AI}  &Java  &  \checkmark& \cite{karakovskiy2012mario}\\
        & & \textit{Morai-Maker-Engine} &C\#  &    & \cite{guzdial2019friend}\\
    &  & \textit{HOG2} & C++  &   & \cite{hog2} \\

    &  & \textit{Procedural Zelda}  &Lua  &   & \cite{heijne2017procedural}\\
    & & \textit{Talakat}  &JavaScript  & & \cite{khalifa2018talakat} \\
        &   & \textit{Keiki}  &Python   &   &\cite{wang2021keiki}\\

     &  & \textit{PCGRL gym} &Python  &   & \cite{khalifa2020pcgrl} \\
         &  & \textit{micro-rct} & Python  &   & \cite{earle2021learning} \\

    \cmidrule{2-6}
    & \multirow{2}{*}{3D level} & \textit{GDMC}  &Python  & \checkmark & \cite{salge2020ai} \\
    &  & \textit{The Sims}  & C++  &     & \cite{charity2020say}\\
\cmidrule{2-6}
    & \multirow{2}{*}{Narrative} &  \textit{TextWorld} & Python & & \cite{cote2019textworld} \\
    & & \textit{Evennia}   & PostScript &     & \cite{ammanabrolu2020bringing}\\
\cmidrule{2-6}
   & \multirow{2}{*}{Rhythm chart}  & StepMania &C++  &     & \cite{StepMania}\\
    &  & \textit{OSU!}  &C\#  &     & \cite{osu}\\
    \cmidrule{2-6}
    & \multirow{2}{*}{Card} &  \textit{XMage} & Java & & \cite{ling2016latent}\\

    & & \textit{Metastone}&Java    &  & \cite{santos2017monte}\\ 
    \midrule
       \multirow{6}{*}{\makecell{\textbf{Other} \\(Section \ref{sec:otherG})}} & Code&\textit{PhyDSL}  & Java   &    & \cite{guana2014phydsl}\\
       \cmidrule{2-6}
       & Persona & \textit{MiniDungeons} & Java   &    & \cite{holmgaard2014evolving}\\

        \cmidrule{2-6}
    & Data set&\textit{VGLC}  & DOT    &    & \cite{James2016vglc}\\

    \cmidrule{2-6}
    & \multirow{2}{*}{Interactive} & \textit{Baba is y'all} & Python & \checkmark & \cite{charity2020baba} \\
    &  & \textit{ Lode Encoder} &JavaScript  &   &  \cite{bhaumik2021lode}\\
    \bottomrule
    \end{tabular}

    \label{tab:designing}
\end{table*}

In recent years, game-based platforms have also been used to create new tasks for AI and to study AI creativity, in particular, the ability of creating new contents. Tab. \ref{tab:designing} and Fig. \ref{fig:gamedesign} categorise the games and platforms used for game design according to research aim including game tuning and content generation, manipulated content type (e.g., parameter, 2D level, 3D level and narrative), programming language and if they have been used by any AI-related competitions. Some resources such as code and dataset are labelled as the other type.
Those platforms provide diverse and representative test-beds for discovering the potentials in AI-involved game design.

\subsection{Game Tuning}
\label{sec:tuning}
Game tuning is vital since it directly affects players' experience. Unsuitable game parameter settings may lead to game unbalance or low quality, where players find it hard to play the game or feel bored.
Generally, all the games introduced in Section \ref{sec:gfp} that involve configurable parameters can be tuned with search-based methods for different purposes, such as adjusting the game difficulty.
For instance, \textit{FightingICE}~\cite{khan2022darefightingice} has been used to study the dynamic difficulty and battle sound adjustment~\cite{ishihara2018monte}. 
\textit{AIGD2}~\cite{lucas2018game}, an open-source game similar to \textit{Planet Wars}, allows to explore the parameter space, which can significantly affect the gameplay via \textit{GVGAI} agents~\cite{liu2017evolving}.
\textit{VGC}~\cite{reis2021vgc} allows adjusting the attributes of \textit{Pokémons}, called meta-game balance. Compared to other platforms, it specifically facilitates the investigation of enabling fair competition in gameplay.
\textit{RaidEnv}~\cite{jeon2023raidenv} aims to optimise boss raid scenarios. Skills and statistics like health and attack can be adjusted by AI techniques, meeting the needs of different situations.

\subsection{Content Generation}

\label{sec:content}

Content generation is involved with applying AI techniques to create new game content. The generated content types include 2D levels, 3D levels, narratives, rhythm charts and cards, following  Liu et al.~\cite{liu2021deep}.
Different content types are represented in different data structures, raising various challenges to content generation research.

\subsubsection{2D Map \& Level}
Game level is the most commonly generated content type. A 2D level is usually presented in a plane, where agents move in two dimensions. 
\textit{Science Birds}~\cite{ferreira2014a} provides a generative platform, which can load XML-based level files to play \textit{Angry Birds}, a physics-based puzzle video game, which requires the level components' physical feasibility and stability. Generating levels for physical-based puzzle games with AI is complex since more physical constraints, larger state space and action space should be considered. Angry Birds AI Level Generation Competition~\cite{renz2019ai} uses \textit{Science Birds} as its simulator.

\begin{figure*}[htbp]
    \centering
\includegraphics[width=\linewidth]{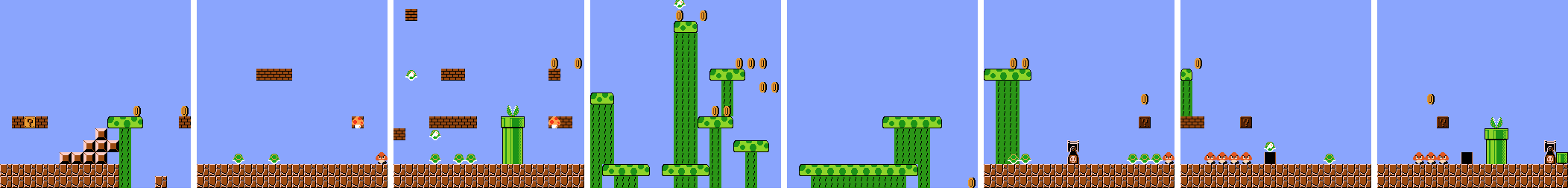}
    \caption{Mario level segments generated by the work of \cite{wang2024negatively} tested and visualised in \textit{Mario AI} framework~\cite{karakovskiy2012mario}.}
    \label{fig:mario_edrl}
\end{figure*}

\textit{Mario-AI-Framework}~\cite{karakovskiy2012mario}, a fully-\texttt{Java} copy of \textit{Super Mario Bros.}, provides a tile-based representation of game levels.
Fig. \ref{fig:mario_edrl} shows examples of Mario levels generated by \cite{wang2024negatively}, tested and visualised in \textit{Mario-AI-Framework}~\cite{karakovskiy2012mario}.
\textit{Morai-Maker}~\cite{guzdial2019friend} is a Unity3D-based platform for \textit{Super Mario Bros.}, which provides three built-in ML-based 2D level generators. Co-creation between humans and AI is supported by the platform with a visible game level editor~\cite{guzdial2019friend}, which distinguishes \textit{Morai-Maker} from other \textit{Super Mario Bros.} platforms. 
\textit{Hierarchical Open Graph 2 (HOG2)}~\cite{hog2} includes several grid-based puzzle games such as \textit{Snakebird}~\cite{sturtevant2020unexpected} for exhaustive PCG~\cite{sturtevant2018exhaustive}.
\textit{Procedural Zelda}~\cite{heijne2017procedural} is a grid-based game platform for level, combat and puzzle generation. Both offline and online control are supported during the generation with data logging capabilities. 

\textit{Talakat}~\cite{khalifa2018talakat} is a bullet hell game engine which defines a domain-specific language for bullet hell levels, supporting the grammar evolution of bullet hell game content.
\textit{Keiki}~\cite{wang2021keiki} is another bullet hell type game engine which provides a native \texttt{Python} interface describing bullet hell barrage. 
While \textit{Talakat} represents barrages as scripts, \textit{Keiki} provides a function to convert barrage scripts into vector-based representations, making it convenient for ML-based content generators.

\textit{Procedural content generation via reinforcement learning (PCGRL) gym}~\cite{khalifa2020pcgrl} implements RL environments for grid-based level generation with the \textit{Gym} interface. Binary, \textit{Zelda} and \textit{Sokoban} are integrated into \textit{PCGRL gym}. 
\textit{micro-rct}~\cite{earle2021learning} is a theme park management simulator, inspired by \textit{RollerCoaster Tycoon 2}. In Earle et al. \cite{earle2021learning}, researchers use a controllable PCGRL to generate game maps. \textit{micro-rct} is used to evaluate the controllability of PCGRL. 
Unlike other platforms which serve for search-based, supervised learning-based, or unsupervised learning-based content generation algorithms, \textit{PCGRL gym}~\cite{earle2021learning} provides a standardised benchmark for RL-based PCG, however, no accessible playing interface for human players is provided.

Recently, LLMs have obtained much attention~\cite{chang2023survey,gallotta2024large}. \textit{Chatgpt4PCG}~\cite{Abdullah2024ChatGPT4PCG} competition aims at using ChatGPT to generate game levels of \textit{Angry Birds}. Although no specific environment API is provided, an example prompt is introduced. \textit{Chatgpt4PCG} enables level generation via natural language, which is more user-friendly for beginners. Meanwhile, the investigation into generating levels with ChatGPT can help researchers to better understand the ability and limitations of LLMs.

\subsubsection{3D Map \& Level}
\begin{figure}[htbp]
    \centering
    \subfigure{
		\centering
		\includegraphics[width=0.45\linewidth]{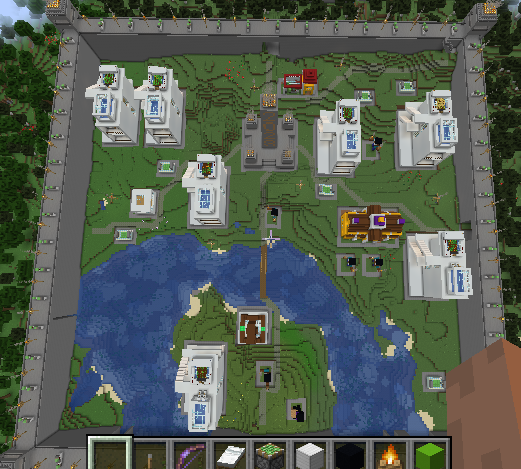}
  }	
\subfigure{
		\centering
		\includegraphics[width=0.43\linewidth]{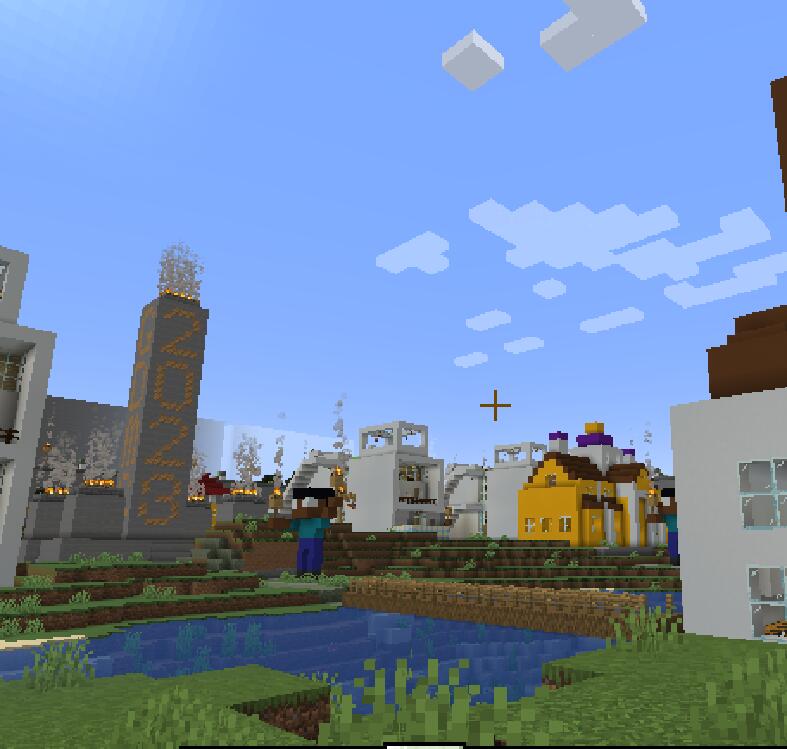}
	}
\caption{\label{fig:gdmc}Screenshots of cities generated by an entry~\cite{huang2023generating} to the GDMC AI Settlement Generation Challenge in \textit{Minecraft}~\cite{salge2018generative} in 2023.}
\end{figure}

3D maps and levels are closer to real-world scenarios but their generation is more complex. Due to the additional dimension, the data scale of 3D maps or levels is typically much larger than those of 2D ones, requiring higher computational efficiency. Moreover, 3D maps and levels remain more complex spatial structures, which lead to higher requirements in playability and diversity.
The \textit{Settlement Generation Competition for \textit{Minecraft} (GDMC)}~\cite{salge2018generative} invites participants to procedurally generate a city or village given a \textit{Minecraft} map, and directly uses the original game engine for simulating and experiencing the output city or village. A code interface is provided to build \textit{Minecraft} maps by setting blocks.
Fig. \ref{fig:gdmc} shows examples of entries to the 2023 edition of GDMC.
\textit{The Sims}~\cite{charity2020say} is a game-inspired platform for indoor environment generation. In \textit{The Sims}, AI techniques can be used to generate furniture and room decorations. 

\subsubsection{Narrative} 
AI techniques can be used to generate interactive fiction worlds with semantic coherence and fun. 
Distinguished from generating other types of content, narrative generation requires AI to understand temporal logic, context memory and natural language.
\textit{WorldGeneration}~\cite{ammanabrolu2020bringing} integrates DL-based and rule-based models to generate graph-based world representations, and then generate narratives of the world. The used game platform is \textit{Evennia}, a \texttt{Python}-based framework for texted-based multiplayer online games, also known as \textit{multi-user dungeon}. \textit{Evennia} has a convenient API and can be easily accessed by the web client. 
\textit{Textworld}~\cite{cote2019textworld} is a text-based engine, which can generate and simulate text games. The aim of \textit{Textworld} is to provide an accessible API for RL agents~\cite{chen2022analysis} to understand natural language and then make decisions.

\subsubsection{Rhythm Chart} Rhythm game is a special 2D game that aims to hit charts at the right time. To optimise player experience, the generated charts need to be synchronous with the rhythm of the music. \textit{StepMania}~\cite{StepMania} is a free dance and rhythm game engine based on game \textit{Dance Dance Revolution} that supports customised rhythm charts. \textit{StepMania} has been researched in level generation, i.e., chart generation~\cite{donahue2017dance}, which is related to the music information retrieval field using long short-term memory (LSTM) and C-LSTM model. \textit{StepMania} can be used in difficulty estimation~\cite{franks2023ordinal} as well. A beatmap and rhythm chart can be generated automatically according to the music~\cite{liang2019procedural,halina2021taikonation} by a rhythm game platform \textit{OSU!}~\cite{osu}.

\subsubsection{Board \& Card} 
Content generation for board and card games encompasses the rules, mechanism or deck building.
\textit{Ludi}~\cite{browne2010evolutionary} is a general game system, aiming at procedurally generating board game rules using evolutionary methods. CCG provides a flexible way to rebuild the game mechanism using AI techniques.
CCGs are challenging in the field of AI due to their vast search space and dynamically changing game states resulting from a large number of diverse cards and their unique functionalities.
The famous CCG, \textit{Hearthstone} has some open-source simulators such as \textit{Metastone}~\cite{santos2017monte} and \textit{Sabberstone}~\cite{dockhorn2019introducing}, which has been used for deck building~\cite{garcia2016evolutionary}, card generation~\cite{ling2016latent} and game playing~\cite{santos2017monte}. \textit{Magic: The Gathering} is another commercial CCG, which has an open-source simulator called \textit{XMage} for AI research including text design~\cite{zilio2018neural} and game playing~\cite{churchill2019magic}.

\subsection{Other Resources for AI-involved Game Design}
\label{sec:otherG}
\textit{PhyDSL} provides an accessible environment to design physics-based 2D games via code generation \cite{guana2014phydsl}. A high-level description language is developed to describe complicated game dynamics in \textit{PhyDSL}.
\textit{MiniDungeons}~\cite{holmgaard2014evolving} is a roguelike dungeon platform for persona modelling, which can be used to build game evaluators.
\textit{Video Game Level Corpus (VGLC)}~\cite{James2016vglc} is an open level set of video games, including \textit{Super Mario Bros.}, \textit{Kid Icarus}, \textit{The legend of Zelda} and \textit{Doom2}, in tile-based, graph-based  (using \texttt{DOT} language in Graphviz~\cite{gansner2000open}) and vector-based representations. 
Although the codes of some platforms are not released, interactive interfaces are accessible to game designers, e.g.,  \textit{Baba is y'all}~\cite{charity2020baba} and \textit{Lode Encoder}~\cite{bhaumik2021lode}.

\subsection{Guidance for Designing Games}

Tab. \ref{tab:designing} summarises the games and platforms for game design, in which how AI can be utilised can be checked in column \textbf{Type}. Programming languages vary across the games and platforms, with half of them supporting \texttt{Java}.
Considering the trend of an increasing portion of using Python in AI research, it is recommended to use those platforms with cross-language programming. For example, researchers can access \texttt{Java} programs in \texttt{Python} via \texttt{Jpype}\footnote{\url{https://github.com/jpype-project/jpype/}}.

Some specific games, such as \textit{FightICE} and \textit{VGC}, can be used for studying game balance and skill discovery via parameter tuning techniques~\cite{reis2023adversarial,khan2023fighting,khan2024enhanced}. 
A majority of games for design focus on procedurally generating content such as 2D and 3D levels. AI techniques including rule-based, search-based, ML-based and DL-based methods have been widely used for PCG using games listed in Tab. \ref{tab:designing}~\cite{togelius2011search,shaker2016procedural,yannakakis2018artificial,Summerville2018Procedural,DeKegel2020Procedural,liu2021deep,guzdial2022procedural,yannakakis2023affective}. For example, bullet hell and maze level can be realised by grammar evolution~\cite{khalifa2018talakat,shaker2012evolving} and quality diversity search~\cite{gravina2018quality}.
Real-time generation of 2D levels of \textit{Super Mario Bros.} represented by latent vectors also benefits from RL~\cite{shu2021experience,wang2024negatively}.
These games also serve as benchmarks for generative models like generative adversarial networks~\cite{volz2018evolving}, diffusion models~\cite{dai2024procedural} and LLMs~\cite{shyam2024mariogpt}. In addition, supervised learning approaches have been applied to rhythm chart generation~\cite{donahue2017dance,halina2021taikonation} using \textit{StepMania} and \textit{OSU!}, as well as card generation~\cite{ling2016latent} based on  \textit{XMage}. 
 Notably, games for content generation like \textit{Procedural Zelda} are also involved with adjusting game difficulty~\cite{gonzalez2020finding}, and tuning parameters of games or agents~\cite{liu2017evolving,fontaine2019mapping,kunanusont2017n}.

Beginners are recommended to try 2D level generation first since it is the most studied with massive platforms. In particular, the \textit{Mario AI} framework is a widely used and representative platform, which is proper to get started. 
\textit{Mario AI}, \textit{Procedural Zelda} and \textit{PCGRL gym} can be used to generate 2D grid-based levels.
Specifically, researchers interested in PCG via RL can use \textit{PCGRL gym}, while the other two platforms are suitable for studying search-based and supervised learning-based level generating methods.
To study the generation of structured game levels with physical property, \textit{Science Birds} is an appropriate choice. While both \textit{Talakat} and \textit{Keiki} are designed for bullet hell games, the former serves for grammar evolution while the latter is convenient for ML research.

For other types of content, there are very limited numbers of platforms to use. Researchers can directly choose the corresponding platform according to the interested content type with the help of Tab. \ref{tab:designing}. \textit{FightingICE}, \textit{AIGD2} and \textit{VGC} are platforms that can be used for studying game tuning and balancing, but with different focus on fighting games, RTS games and \textit{Pokémon}, respectively.

\section{Discussion and Outlook}\label{sec:discuss}
 New platforms and games have been published over the last few years as ideal test-beds for AI research, including multi-agent RL, artificial general intelligence and creative AI. This section discusses the core challenges from games, our key findings and outlook.

\subsection{Challenges from Games}

Although originally designed for human entertainment, games have proven to be ideal test-beds for validating the capabilities of AI techniques, providing a range of challenges for AI to overcome.
\subsubsection{Curse of High Dimensions}
Dimensions of games are involved with the number of game states, representation complexity of state and action spaces. For example, \textit{Go} has about $3^{361}$ possible positions, encoded as a 0-1 matrix. Modern games that rely on digital devices have a more sophisticated, high-dimensional representation, i.e., video frames. Traditional tabular methods such as Q-learning~\cite{watkins1992q} fail to deal with the situation since the required space scales exponentially with the dimension. It is from deep Q-learning~\cite{mnih2015human} that the neural network appears on the scenes as the direct state estimator. In recent years, the development of games has resulted in fancy game screens with high-quality graphics, which can immerse players in more sophisticated virtual worlds such as \textit{Atari}~\cite{brockman2016openai}, \textit{Dota2}~\cite{berner2019dota} and \textit{StarCraft II}~\cite{vinyals2019grandmaster}. However, this visual sophistication makes it hard, particularly for AI agents to extract relevant features from those games~\cite{berner2019dota}. Well-rendered sprites can obscure critical information that AI agents need to make decisions. For example, in a game where players control characters that move around in an environment, the sprites can prevent AI agents from identifying obstacles and enemies. Furthermore, video games often provide players with a large and complex set of actions, leading to a combinatorial explosion of possible actions~\cite{dulac2015deep}. 

\subsubsection{Poor Exploration}
AI techniques such as RL usually need a large amount of interaction data to train policies. It is hard to directly access such a ``dataset''. Instead, we need to collect the data, denoted as experience, during training, which leads to a tradeoff between exploration and exploitation. The tradeoff happens with a limited computational budget since we hardly go through all possible states and actions. \textit{Montezuma's revenge} epitomises the games where AI techniques suffer from the tradeoff~\cite{burda2018exploration}, which brings a deep thought about the exploration strategy. Integrating the exploration signal into the reward function has been shown efficient to tackle the challenge, such as count-based exploration~\cite{bellemare2016unifying}, intrinsic curiosity~\cite{burda2018largescale} and novelty search~\cite{conti2018improving}. While the above reward shaping may lead to the optimal policy variance, population-based exploration~\cite{majid2023deep} and noisy exploration~\cite{burda2018exploration,fortunato2018noisy} that perturb the policy parameter space, present their advantage for the diverse exploration.

\subsubsection{Partial Observability}
While complete information is known in \textit{Atari}~\cite{brockman2016openai} and \textit{Go}~\cite{silver2017mastering} at a given state, partial observability is encountered in more complex video games such as \textit{ViZDoom}~\cite{kempka2016vizdoom}. There are two meanings associated with partial observability: agents cannot obtain perfect information about the current state, and agents cannot determine the tendencies of other agents. The latter usually refers to the multi-agent case. In the former aspect, agents only can partially observe the environment~\cite{jaakkola1994reinforcement}. \textit{ViZDoom}~\cite{kempka2016vizdoom} only allows players to view the video frame from a first-person perspective. Although agents can handle situations in front of them, unexpected occurrences such as monsters approaching from behind should also be considered.

\subsubsection{Generalisation}
Although AI techniques such as deep RL have achieved a competitive level beyond humans in games including \textit{Go}~\cite{silver2017mastering} and \textit{Atari}~\cite{brockman2016openai}, their generalisation ability is questionable since they are usually only trained on a single game or level. The generalisation considers if the agent can still play well in some unseen games and levels without overfitting the training environment~\cite{kirk2023survey}. Many game-based environments are involved with the topic. \textit{GVGAI}~\cite{perez2016general,perez2019general} focus on the performance of agents when trained on some game levels but tested on unseen game levels. The unseen test game levels have different game maps or layouts, but their mechanism is the same \cite{cobbe2020leveraging}. Both planning and learning algorithms are supported by the \textit{GVGAI} environment~\cite{perez2016general,perez2019general}.
\textit{Procgen}~\cite{cobbe2020leveraging} also provides diverse game scenarios, which test the AI agents by randomly changing the initial position of the avatars.

\subsubsection{Multi-agent Games}
It is common to see the participation of multiple agents or players in games such as tabletop games \cite{kowalski2023introducing,fachada2021colorshapelinks}, multiplayer online battle arena games~\cite{berner2019dota}, and RTS games~\cite{vinyals2019grandmaster}. Three main multi-agent settings are involved: fully cooperative, fully competitive, and mixed. For example, in \textit{Dota2}, five players are matched as a team to fight against another 5-player team. The five players cooperate and share their resources and experience, while also competing with the other team. This participation leads to an increase in dimension with the expanding number of agents. Additionally, non-stationarity and credit assignment problems arise when multiple independent agents exist since it is difficult to determine the unique contribution of each agent. Another interesting exploration is the coordination between AI agents and humans. While humans typically rely on natural language to communicate and cooperate, AI agents may not have the same capacity. The recent emergence of LLMs~\cite{zhao2023survey} such as ChatGPT~\cite{openai2023chatgpt} shows the potential for human-machine interaction. The work of~\cite{xu2023exploring} present how LLMs cooperate in 7-person \textit{Werewolf} without tuning. LLM-based players collect experience in each game round and demonstrate strategic behaviours. It remains to be seen if this success can be replicated in games such as \textit{Hanabi}~\cite{bard2020hanabi} by both communicating and acting. Integrating LLMs with current AI technologies such as RL is a potential solution. In this combination, LLMs can understand the current game state and extract information, which enables an RL agent to make reasonable decisions.

\subsection{AI for Games}

The traditional developing process of games is long and complex, which burdens humans in designing and programming game content.
Nowadays, AI can assist humans to improve games by testing and generating new content.

\subsubsection{Game Testing by AI}
Game quality assurance is crucial in the gaming industry for assessing game content~\cite{politowski2021survey}. However, employing human testers incurs significant financial and time costs. To solve this problem, automated game testing has been widely studied for decades. With AI technologies such as search-based methods~\cite{fraser2011evosuite} and RL~\cite{zheng2019wuji,bergdahl2020augmenting}, bugs in games can be automatically detected and fixed. Yet, current AI testing methods mainly lack sufficient generalisation across different games. Typically, AI-based testing solutions require additional costs of being adapted to new games. For example, search-based methods need to search for new test cases and learning-based methods require retraining~\cite{zheng2019wuji}.
Developing a more generic testing framework with lower cost is invaluable. A promising direction may involve transfer learning~\cite{pan2009survey}, which enables the automated testing framework to adapt to new games, making high-quality assurance more accessible for developers across the game industry.

Moreover, games often require frequent updates to introduce new content and address gameplay issues. This necessity can be both a source of excitement and frustration for players. The critical challenge lies in testing updated content and maintaining game balance. Such studies~\cite{preuss2018integrated,reis2023adversarial} have been conducted on multiple games such as \textit{VGC}~\cite{reis2021vgc} and  \textit{StarCraft II}~\cite{vinyals2019grandmaster}. Advanced testing methodologies are capable of adapting to the rapidly changing landscape of game development, which enhances the player experience without compromising game quality or balance.

 \subsubsection{Game Design by AI}
In games, levels is usually the most fantastic part, which represents the effort and inspiration of game designers. However, creating levels requires a significant investment of time and human resources.
AI has the potential of accelerating this process and can even serve as a substitute for human designers~\cite{togelius2011search,shaker2016procedural,DeKegel2020Procedural,liu2021deep,guzdial2022procedural}.

Besides, AI can be applied to generate music and images in games. For example, Ferreira et al~\cite{ferreira2020computer}. proposed a DL-based music composition system to generate music for tabletop role-playing games. Hong et al. applied multi-discriminator generative adversarial networks to generate game sprites \cite{hong2019game}.  On the other hand, other 2D and 3D content generation methods~\cite{liu2024comprehensive}, not limited to AI approaches, can also be tested and applied to games.
Moreover, game stories and programming codes can be generated via AI technologies. For example, Ubisoft has recently proposed \textit{Ghostwriter} to assist game designers to write game stories and dialogues~\cite{barth2023ghostwriter}. \textit{Ghostwriter} receives feedbacks from designers to construct conversations between NPCs, instead of replacing designers directly.

Given these advancements in AI technologies, an interesting question arises: is it possible for AI to generate a complete game? \textit{GameGAN} proposed by NVIDIA~\cite{kim2020learning} presents this possibility. After training with massive playing records in \textit{Pac-Man}, \textit{GameGAN} can simulate the game without relying on a game engine or coding. When the player pushes some buttons, \textit{GameGAN} predicts the next game frame for the player.  More recently, DeepMind has proposed \textit{Genie}~\cite{bruce2024genie}, a generative interactive environment directly trained by videos from Internet. Beyond \textit{GameGAN}~\cite{kim2020learning}, \textit{Genie} generates endless virtual worlds in a more controllable way by inputting prompts and photos, with which players can directly interact. However, \textit{Genie} is limited to generating 2D games presently. Besides, the generated game frames can be hardly modified manually.
It is believed that there will be more AIs capable of generating complete games, marking the dawn of a new era in game development.

\subsection{Games for AI}
Beside the specific challenges such as high dimensions and partial observability, games can enhance AI by providing unique testing environments. Games not only expand AI capabilities, but also are regarded as preliminary experiments before applying AI to real-world scenarios.

 \subsubsection{Open-world Games for Artificial General Intelligence}
 General (video) game playing has been extensively studied using multi-game platforms~\cite{genesereth2005general,perez2016general,perez2019general,liebana2020general}. However, the objectives or tasks in such games are often well-defined and known. 
 Open-world games, e.g., \textit{MineRL}~\cite{guss2019minerldata}, pose new challenges for AI to autonomously explore the \textit{Minecraft} world without specific aims, including the requirement of accomplishing a series or set of tasks instead of one or two, with inefficient data sampling and indeterminate reward function~\cite{guss2019minerldata}. The nature of open-world games provides an ideal platform to develop and evaluate artificial general intelligence. They can be considered as sandboxes, where AI agents are not only tested on some specific tasks such as navigating and building, but also are required to interact with the dynamic environment. The diverse tasks such as farming, hunting and crafting in \textit{Minecraft} enable AI to live as humans.
This case is pushing the boundaries of AI towards achieving artificial general intelligence, as it mimics the complexity and unexpected cases of real-world tasks and decision-making.

\subsubsection{Games as Reflection of Real-world Scenarios}
Games can simulate real-world scenarios and enable us to make new attempts of AI safely at a low cost and reflect challenges in real-world scenarios. For example, 
urban planning can be found in games such as \textit{SimCity}~\cite{earle2020using}. 
Building and sculpture design prototypes are found in \textit{GDMC}~\cite{salge2020ai}. \textit{AI wolf}~\cite{toriumi2016ai} benefits for understanding the behaviours of AI in cooperative and competitive settings.
Logistics problems are challenging in the real world and it can be difficult to evaluate the real-time decision algorithms directly in real-world manufacturing systems due to some safety and property constraints. These problems can be easily modelled in games to validate the effectiveness and safety of algorithms. For example, \textit{Shapez}~\footnote{\url{https://github.com/tobspr-games/shapez.io}} is a building game that aims to simulate manufacturing by placing and combining different belts and workstations. AI can test extensive possible solutions with the game at a low cost, which contributes to the design of a factory in the real world. It is notable there are also many games like \textit{Mini Metro}, \textit{Mini Motorway} and \textit{Cities: Skylines} that reflect real-world scenarios. However, the codes of those commercial games are not released.

 \subsubsection{Emerging Environments for AI Creativity}
The environments collected in this paper indicate the research trend from intelligent decision-making to creative AI. Since 2016, 14 new environments have been published for game design by AI. However, most of them aim at designing tile-based levels. In addition, the collected environments are often not equipped with detailed descriptions or documentation as they were released as dependencies of a published algorithm or ML model.
 Although PCG approaches have been studied or applied to a wide range of games, the environments are rarely open-source (e.g., \textit{Cut the Rope}~\cite{shaker2013ropossum}). Besides, how to design effective content-evaluation metrics to enable promising content generation is a non-trivial problem.
 More open-source environments with diverse genres and content types, user-friendly APIs and documents, are expected to advance the research in AI creativity and human-AI co-creativity.

 \subsection{Under-explored Game Genres}
 
 Although there are many available games, they share a large number of similar mechanisms with only differences in sprite styles. Many platforms for AI research are based on classic games like \textit{Atari}~\cite{brockman2016openai}. Although the simplified representation diminishes some engineering problems, it is hard to construct more delicate game mechanisms. Only a few modern commercial games such as \textit{StarCraft II}~\cite{vinyals2019grandmaster}, and \textit{Dota 2}~\cite{berner2019dota}, provide accessible APIs or lightweight simulators.
 
Extensive effort has to be put into solving engineering problems such as controlling avatars, which is a critical issue in conducting AI research on commercial games. A possible solution for this is capturing the screening as input. AI agent directly accesses the screen frames and plays the game via a virtual keyboard and mouse.  As new games are developed, challenges arise for AI research. It is expected that more commercial games provide accessible APIs that AI can interact with.
 Similar concerns exist in game-based platforms as most platforms for content generation are based on rather classic video games like \textit{Super Mario Bros.}~\cite{karakovskiy2012mario}, \textit{Angry Birds}~\cite{ferreira2014a,renz2019ai} and \textit{Minecraft}~\cite{guss2019minerldata,fan2022minedojo}. Though using classical games makes it convenient for research, it prevents AI-involved game design from being used in practice.

\subsection{Lower Barriers for Building New Games}

There are some professional game engines to build simulators for ML research such as \textit{Unity}~\cite{juliani2018unity}, \textit{Unreal Engine}\footnote{\url{https://github.com/zfw1226/gym-unrealcv}} and \textit{Godot}~\cite{beeching2021godot}. 
Developing a domain-specific platform for a game is not easy. Some simulators can only be used in specific scenarios, and if the AI algorithm needs to be used in other situations, building a new simulator may be necessary, which is time-consuming. One way to make simulators more versatile is by applying description languages. \textit{VGDL}~\cite{schaul2013pyvgdl} is a good example. Games written by \textit{VGDL} can be compiled by a specific compiler such as \textit{GVGAI}~\cite{perez2016general,perez2019general,liebana2020general} and \textit{UnityVGDL}~\cite{johansen2019video}. Another way is using general platforms, especially professional game engines to build simulators such as \textit{Unity}~\cite{juliani2018unity}, \textit{Unreal Engine} and \textit{Godot}~\cite{beeching2021godot}. Using professional game engines, simulators can be built much easier and run faster. With remote procedure call, algorithms do not need to be implemented with the same programming language as simulators do, which makes it possible to use general platforms for AI research, reducing engineering issues. 

\subsection{Large Language Models for Games}

Recently, the emergence of LLMs such as ChatGPT~\cite{openai2023chatgpt} narrows the gap of game generation~\cite{gallotta2024large}. Using natural languages, i.e., interacting with LLMs, everyone can create their own games. Some other applications of LLMs for assisting game development such as level generation and dialogue generation also thrive. MarioGPT~\cite{Sudhakaran2023prompt,shyam2024mariogpt} provides an end-to-end Mario level generator through a fine-tuned GPT2. Users can write some simple prompts such as ``many pipes, some enemies'' to generate game levels with their preference. The work of \cite{Todd2023level} validates the effectiveness of LLMs for generating \textit{Sokoban} levels from the perspective of playability, novelty, diversity and accuracy.

Instead of assisting game content generation, LLMs themselves can also be regarded as a part of games. NPCs have been playing an essential role in games. Traditionally, NPCs have been confined to fixed scripts, limiting them to repetitive conversations that do not interact with or affect players' strategies. LLMs, however, provide the potential of human-like NPCs, who may act and respond according to the behaviours of players. For example, Park et al.~\cite{park2023generative} adopt ChatGPT for the generative agent design. Players directly communicate with NPCs controlled by ChatGPT. Kumaran et al.~\cite{Kumaran2023scenecraft} propose SceneCraft based on LLMs for narrative generation in a 3D game. In addition to dialogue, SceneCraft~\cite{Kumaran2023scenecraft} generates emotion and gestures of NPCs during the interactions with human players. The work of \cite{qian2023communicative} shows the possibility of developing software including games using multiple LLM agents.

Notably, LLMs also show the potential of playing games. Wu et al.~\cite{wu2024spring} use GPT-4 to read a paper about game \textit{Spring}, and then directly apply GPT-4 to play the game. According to \cite{wu2024spring}, GPT-4 presents an outstanding performance than advanced RL algorithms without training. It is widely believed that LLMs, with their remarkable capabilities and capacity for generalisation, are driving the future direction of games and game AI development.

\subsection{Roles of Games in Interdisciplinary Research in AI}
Games not only play an important role in the development of AI, but also is an ideal tool for studying interdisciplinary research in AI and other computing-related areas, such as social intelligence of AI~\cite{puig2021watchandhelp}, affective computing~\cite{yannakakis2023affective}, health and education~\cite{Bartolomé2011can,li2013game,abiyev2016brain}, cognitive behavioural therapy for children~\cite{brezinka2014computer}. Besides, game-based learning has presented its strong capability in developing problem-solving and cooperative skills of children, compared with traditional lessons~\cite{liu2020using}. Games have also been used as the mediums for educating AI~\cite{zhao2024playing}.

\section{Conclusion}\label{sec:con}

This paper reviews and categorises the publicly available games and game-based platforms for AI research. Growing research interests in AI for game testing, and AI for game design, games for artificial general intelligence, multi-agent systems, and games for creative AI, have been induced. We present the games and game-based platforms for playing according to their characteristics, such as aims, the number of agents, observability, programming languages, and whether AI related competitions were held. Games and platforms for design are grouped by focused content type, such as 2D levels, 3D levels, narratives, rhythm charts, and cards. 

Games serve as controllable and customisable simulators. Consequently, we discuss existing challenges thriving from games. The rapid development of games leads to the curse of high dimensions. Poor exploration and partial observability are also commonly found in games since agents need to interact with the environment. Generalisation capability of AI agents is validated on games, which poses the question if they still work in some unseen games and levels. 
The multi-agent games, where non-stationarity, coordination and human-AI interaction are involved, are also discussed.

Nevertheless, this paper presents how AI works for games and games work for AI. As a perfect assistant to humans, AI is capable of testing games and balancing game content during the process of game development. Besides, AI can be applied to generate different types of game content, such as levels, music and stories, which highly releases human resources. Open-world games provide virtual worlds for artificial general intelligence to explore. As the reflection of real-world scenarios, games can be applied as preliminary test cases.
Furthermore, the recent emergence of LLMs offers a new perspective on the development of AI techniques. Natural languages can be directly used for creative design, such as levels, images and the games themselves. Recent studies present how LLMs make actions in games according to the textual representations. LLMs also show the potential of directly engaging in gameplay as NPCs, which entertains human players.

However, some concerns about the development of game AI platforms are raised. 
Although AI works for game testing and design, it lacks a united framework across games. Essential efforts have to be put in when applying AI to testing new games or designing game content.
Available interfaces for modern commercial games are limited by some engineering issues. A large number of commercial games do not provide official APIs for AI research.
Although existing platforms offer a controllable, safe and diverse environment to study, explore, evaluate and experiment with different ideas, the augmentation of test-beds for AI would be facilitated by more accessible interfaces for building new games. 
Considering the gaps between academics and industries that own plenty of computational resources and copyright of commercial games~\cite{togelius2024choose}, getting started for beginners in game AI on some simple environments such as \textit{Atari} may be a good choice.
It is expected to see new open-source game-based platforms with under-explored game genres, diverse content types, user-friendly APIs and documents.

Finally, new, open-ended questions arise while this paper was being written and, at the same time, LLMs are rapidly advancing almost with daily improvements: (i) What challenges that motivated the implementation of the games/platforms reviewed by this paper or were posed by those games/platforms cannot be solved by LLMs in the near future and remain valuable to study?  (ii) What the games/platforms reviewed by this paper can contribute to the development of LLMs?

\section*{Acknowledgement}
Authors would like to thank all anonymous reviewers for their careful review and insightful comments.

\bibliographystyle{IEEEtran}

\bibliography{main}

\begin{thebibliography}{100}
\providecommand{\url}[1]{#1}
\csname url@samestyle\endcsname
\providecommand{\newblock}{\relax}
\providecommand{\bibinfo}[2]{#2}
\providecommand{\BIBentrySTDinterwordspacing}{\spaceskip=0pt\relax}
\providecommand{\BIBentryALTinterwordstretchfactor}{4}
\providecommand{\BIBentryALTinterwordspacing}{\spaceskip=\fontdimen2\font plus
\BIBentryALTinterwordstretchfactor\fontdimen3\font minus \fontdimen4\font\relax}
\providecommand{\BIBforeignlanguage}[2]{{%
\expandafter\ifx\csname l@#1\endcsname\relax
\typeout{** WARNING: IEEEtran.bst: No hyphenation pattern has been}%
\typeout{** loaded for the language `#1'. Using the pattern for}%
\typeout{** the default language instead.}%
\else
\language=\csname l@#1\endcsname
\fi
#2}}
\providecommand{\BIBdecl}{\relax}
\BIBdecl

\bibitem{swiechowski2020game}
M.~{\'S}wiechowski, ``Game {AI} competitions: Motivation for the imitation game-playing competition,'' in \emph{Conference on Computer Science and Information Systems}.\hskip 1em plus 0.5em minus 0.4em\relax IEEE, 2020, pp. 155--160.

\bibitem{mnih2015human}
V.~Mnih, K.~Kavukcuoglu, D.~Silver, A.~A. Rusu, J.~Veness, M.~G. Bellemare, A.~Graves, M.~Riedmiller, A.~K. Fidjeland, G.~Ostrovski, S.~Petersen, C.~Beattie, A.~Sadik, I.~Antonoglou, H.~King, D.~Kumaran, D.~Wierstra, S.~Legg, and D.~Hassabis, ``Human-level control through deep reinforcement learning,'' \emph{Nature}, vol. 518, no. 7540, pp. 529--533, 2015.

\bibitem{van2016deep}
H.~Van~Hasselt, A.~Guez, and D.~Silver, ``Deep reinforcement learning with double {Q}-learning,'' in \emph{Proceedings of the AAAI Conference on Artificial Intelligence}, vol.~30, no.~1, 2016.

\bibitem{chinchali2018cellular}
S.~Chinchali, P.~Hu, T.~Chu, M.~Sharma, M.~Bansal, R.~Misra, M.~Pavone, and S.~Katti, ``Cellular network traffic scheduling with deep reinforcement learning,'' in \emph{Proceedings of the AAAI Conference on Artificial Intelligence}, vol.~32, no.~1, 2018.

\bibitem{yannakakis2018artificial}
G.~N. Yannakakis and J.~Togelius, \emph{Artificial Intelligence and Games}.\hskip 1em plus 0.5em minus 0.4em\relax Springer, 2018, vol.~2.

\bibitem{liu2021deep}
J.~Liu, S.~Snodgrass, A.~Khalifa, S.~Risi, G.~N. Yannakakis, and J.~Togelius, ``Deep learning for procedural content generation,'' \emph{Neural Computing and Applications}, vol.~33, no.~1, pp. 19--37, 2021.

\bibitem{de2018games}
S.~De~Freitas, ``Are games effective learning tools? a review of educational games,'' \emph{Journal of Educational Technology \& Society}, vol.~21, no.~2, pp. 74--84, 2018.

\bibitem{levine2013general}
J.~Levine, C.~B. Congdon, M.~Ebner, G.~Kendall, S.~M. Lucas, R.~Miikkulainen, T.~Schaul, and T.~Thompson, ``{General Video Game Playing},'' in \emph{Artificial and Computational Intelligence in Games}.\hskip 1em plus 0.5em minus 0.4em\relax Schloss Dagstuhl -- Leibniz-Zentrum f{\"u}r Informatik, 2013, vol.~6, pp. 77--83.

\bibitem{oikarinen2021robust}
T.~Oikarinen, W.~Zhang, A.~Megretski, L.~Daniel, and T.-W. Weng, ``Robust deep reinforcement learning through adversarial loss,'' \emph{Advances in Neural Information Processing Systems}, vol.~34, pp. 26\,156--26\,167, 2021.

\bibitem{ji2023safety}
J.~Ji, B.~Zhang, J.~Zhou, X.~Pan, W.~Huang, R.~Sun, Y.~Geng, Y.~Zhong, J.~Dai, and Y.~Yang, ``Safety gymnasium: A unified safe reinforcement learning benchmark,'' \emph{Advances in Neural Information Processing Systems}, vol.~36, 2023.

\bibitem{spronck2020artificial}
P.~Spronck, J.~Liu, T.~Schaul, and J.~Togelius, \emph{Artificial and Computational Intelligence in Games: Revolutions in Computational Game {AI}: Report from Dagstuhl Seminar 19511}.\hskip 1em plus 0.5em minus 0.4em\relax Schloss Dagstuhl -- Leibniz-Zentrum f{\"u}r Informatik, 2020.

\bibitem{campbell2002deep}
M.~Campbell, A.~J. Hoane~Jr, and F.-h. Hsu, ``Deep blue,'' \emph{Artificial Intelligence}, vol. 134, no. 1-2, pp. 57--83, 2002.

\bibitem{bowling2015heads}
M.~Bowling, N.~Burch, M.~Johanson, and O.~Tammelin, ``Heads-up limit hold'em poker is solved,'' \emph{Science}, vol. 347, no. 6218, pp. 145--149, 2015.

\bibitem{silver2017mastering}
D.~Silver, J.~Schrittwieser, K.~Simonyan, I.~Antonoglou, A.~Huang, A.~Guez, T.~Hubert, L.~Baker, M.~Lai, A.~Bolton, Y.~Chen, T.~Lillicrap, F.~Hui, L.~Sifre, G.~van~den Driessche, T.~Graepel, and D.~Hassabis, ``Mastering the game of {Go} without human knowledge,'' \emph{Nature}, vol. 550, no. 7676, pp. 354--359, 2017.

\bibitem{vinyals2019grandmaster}
O.~Vinyals, I.~Babuschkin, W.~M. Czarnecki, M.~M. andAndrew Dudzik, J.~Chung, D.~H. Choi, R.~Powell, T.~Ewalds, P.~Georgiev, J.~Oh, D.~Horgan, M.~Kroiss, I.~Danihelka, A.~Huang, L.~Sifre, T.~Cai, J.~P. Agapiou, M.~Jaderberg, A.~S. Vezhnevets, R.~Leblond, T.~Pohlen, V.~Dalibard, D.~Budden, Y.~Sulsky, J.~Molloy, T.~L. Paine, C.~Gulcehre, Z.~Wang, T.~Pfaff, Y.~Wu, R.~Ring, D.~Yogatama, D.~Wünsch, K.~McKinney, O.~Smith, T.~Schaul, T.~Lillicrap, K.~Kavukcuoglu, D.~Hassabis, C.~Apps, and D.~Silver, ``Grandmaster level in {StarCraft} {II} using multi-agent reinforcement learning,'' \emph{Nature}, vol. 575, no. 7782, pp. 350--354, 2019.

\bibitem{berner2019dota}
C.~Berner, G.~Brockman, B.~Chan, V.~Cheung, P.~D{\k{e}}biak, C.~Dennison, D.~Farhi, Q.~Fischer, S.~Hashme, C.~Hesse \emph{et~al.}, ``Dota 2 with large scale deep reinforcement learning,'' \emph{arXiv preprint arXiv:1912.06680}, 2019.

\bibitem{pmlr-v139-zha21a}
D.~Zha, J.~Xie, W.~Ma, S.~Zhang, X.~Lian, X.~Hu, and J.~Liu, ``Douzero: Mastering {DouDizhu} with self-play deep reinforcement learning,'' in \emph{International Conference on Machine Learning}, vol. 139.\hskip 1em plus 0.5em minus 0.4em\relax PMLR, 2021, pp. 12\,333--12\,344.

\bibitem{zhao2022alphaholdem}
E.~Zhao, R.~Yan, J.~Li, K.~Li, and J.~Xing, ``{AlphaHoldem}: {H}igh-performance artificial intelligence for heads-up no-limit poker via end-to-end reinforcement learning,'' in \emph{Proceedings of the AAAI Conference on Artificial Intelligence}, vol.~36, no.~4, 2022, pp. 4689--4697.

\bibitem{lu2023danzero}
Y.~Lu, Y.~Zhao, W.~Zhou, H.~Li \emph{et~al.}, ``Danzero: Mastering {GuanDan} game with reinforcement learning,'' in \emph{2023 IEEE Conference on Games (CoG)}.\hskip 1em plus 0.5em minus 0.4em\relax IEEE, 2023, pp. 1--8.

\bibitem{schnier2021nature}
T.~Schnier, R.~Beale, X.~Yao, B.~Hendley, and W.~Byrne, ``Nature inspired creative design--bringing together ideas from nature, computer science, engineering, art and design,'' in \emph{Designing for the 21st Century}.\hskip 1em plus 0.5em minus 0.4em\relax Routledge, 2021, pp. 192--204.

\bibitem{risi2020increasing}
S.~Risi and J.~Togelius, ``Increasing generality in machine learning through procedural content generation,'' \emph{Nature Machine Intelligence}, vol.~2, no.~8, pp. 428--436, 2020.

\bibitem{togelius2011search}
J.~Togelius, G.~N. Yannakakis, K.~O. Stanley, and C.~Browne, ``Search-based procedural content generation: A taxonomy and survey,'' \emph{IEEE Transactions on Computational Intelligence and AI in Games}, vol.~3, no.~3, pp. 172--186, 2011.

\bibitem{shaker2016procedural}
N.~Shaker, J.~Togelius, and M.~J. Nelson, \emph{Procedural Content Generation in Games}.\hskip 1em plus 0.5em minus 0.4em\relax Springer, 2016.

\bibitem{Summerville2018Procedural}
A.~Summerville, S.~Snodgrass, M.~Guzdial, C.~Holmg{\aa}rd, A.~K. Hoover, A.~Isaksen, A.~Nealen, and J.~Togelius, ``Procedural content generation via machine learning {(PCGML)},'' \emph{IEEE Transactions on Games}, vol.~10, no.~3, pp. 257--270, 2018.

\bibitem{DeKegel2020Procedural}
B.~De~Kegel and M.~Haahr, ``Procedural puzzle generation: A survey,'' \emph{IEEE Transactions on Games}, vol.~12, no.~1, pp. 21--40, 2020.

\bibitem{guzdial2022procedural}
M.~Guzdial, S.~Snodgrass, and A.~J. Summerville, \emph{Procedural content generation via machine learning: An Overview}.\hskip 1em plus 0.5em minus 0.4em\relax Springer, 2022.

\bibitem{yannakakis2023affective}
G.~N. Yannakakis and D.~Melhart, ``Affective game computing: A survey,'' \emph{Proceedings of the IEEE}, vol. 111, no.~10, pp. 1423--1444, 2023.

\bibitem{park2023generative}
J.~S. Park, J.~O'Brien, C.~J. Cai, M.~R. Morris, P.~Liang, and M.~S. Bernstein, ``Generative agents: Interactive simulacra of human behavior,'' in \emph{36th Annual ACM Symposium on User Interface Software and Technology}, 2023, pp. 1--22.

\bibitem{Kumaran2023scenecraft}
V.~Kumaran, J.~Rowe, B.~Mott, and J.~Lester, ``{SceneCraft}: Automating interactive narrative scene generation in digital games with large language models,'' \emph{Proceedings of the AAAI Conference on Artificial Intelligence and Interactive Digital Entertainment}, vol.~19, no.~1, pp. 86--96, 2023.

\bibitem{Todd2023level}
G.~Todd, S.~Earle, M.~U. Nasir, M.~C. Green, and J.~Togelius, ``Level generation through large language models,'' in \emph{International Conference on the Foundations of Digital Games}.\hskip 1em plus 0.5em minus 0.4em\relax ACM, 2023, pp. 1--8.

\bibitem{Sudhakaran2023prompt}
S.~Sudhakaran, M.~Gonz\'{a}lez-Duque, C.~Glanois, M.~Freiberger, E.~Najarro, and S.~Risi, ``Prompt-guided level generation,'' in \emph{Proceedings of the Companion Conference on Genetic and Evolutionary Computation}.\hskip 1em plus 0.5em minus 0.4em\relax ACM, 2023, pp. 179--182.

\bibitem{shyam2024mariogpt}
S.~Sudhakaran, M.~Gonz{\'a}lez-Duque, M.~Freiberger, C.~Glanois, E.~Najarro, and S.~Risi, ``{MarioGPT}: {O}pen-ended text2level generation through large language models,'' \emph{Advances in Neural Information Processing Systems}, vol.~36, pp. 1--15, 2024.

\bibitem{wu2024spring}
Y.~Wu, S.~Y. Min, S.~Prabhumoye, Y.~Bisk, R.~R. Salakhutdinov, A.~Azaria, T.~M. Mitchell, and Y.~Li, ``Spring: Studying papers and reasoning to play games,'' in \emph{Advances in Neural Information Processing Systems}, vol.~36, 2023, pp. 1--13.

\bibitem{bruce2024genie}
J.~Bruce, M.~Dennis, A.~Edwards, J.~Parker-Holder, Y.~Shi, E.~Hughes, M.~Lai, A.~Mavalankar, R.~Steigerwald, C.~Apps, Y.~Aytar, S.~Bechtle, F.~Behbahani, S.~Chan, N.~Heess, L.~Gonzalez, S.~Osindero, S.~Ozair, S.~Reed, J.~Zhang, K.~Zolna, J.~Clune, N.~de~Freitas, S.~Singh, and T.~Rocktäschel, ``Genie: Generative interactive environments,'' \emph{arXiv preprint: arXiv: 2402.15391}, 2024.

\bibitem{shao2019survey}
K.~Shao, Z.~Tang, Y.~Zhu, N.~Li, and D.~Zhao, ``A survey of deep reinforcement learning in video games,'' \emph{arXiv preprint arXiv:1912.10944}, 2019.

\bibitem{giannakos2020games}
M.~Giannakos, I.~Voulgari, S.~Papavlasopoulou, Z.~Papamitsiou, and G.~Yannakakis, ``Games for artificial intelligence and machine learning education: Review and perspectives,'' \emph{Non-Formal and Informal Science Learning in the ICT Era}, pp. 117--133, 2020.

\bibitem{duan2022survey}
J.~Duan, S.~Yu, H.~L. Tan, H.~Zhu, and C.~Tan, ``A survey of embodied {AI}: From simulators to research tasks,'' \emph{IEEE Transactions on Emerging Topics in Computational Intelligence}, vol.~6, no.~2, pp. 230--244, 2022.

\bibitem{ibarz2021train}
J.~Ibarz, J.~Tan, C.~Finn, M.~Kalakrishnan, P.~Pastor, and S.~Levine, ``How to train your robot with deep reinforcement learning: lessons we have learned,'' \emph{The International Journal of Robotics Research}, vol.~40, no. 4-5, pp. 698--721, 2021.

\bibitem{kiran2021deep}
B.~R. Kiran, I.~Sobh, V.~Talpaert, P.~Mannion, A.~A. Al~Sallab, S.~Yogamani, and P.~P{\'e}rez, ``Deep reinforcement learning for autonomous driving: A survey,'' \emph{IEEE Transactions on Intelligent Transportation Systems}, vol.~23, no.~6, pp. 4909--4926, 2021.

\bibitem{dosovitskiy2017carla}
A.~Dosovitskiy, G.~Ros, F.~Codevilla, A.~Lopez, and V.~Koltun, ``{CARLA}: An open urban driving simulator,'' in \emph{Conference on Robot Learning}.\hskip 1em plus 0.5em minus 0.4em\relax PMLR, 2017, pp. 1--16.

\bibitem{Nvidia2021Omniverse}
Nvidia, ``Nvidia omniverse,'' \url{https://www.nvidia.com/en-us/omniverse/}, last accessed on 19 April 2024.

\bibitem{todorov2012mujoco}
E.~Todorov, T.~Erez, and Y.~Tassa, ``{MuJoCo}: A physics engine for model-based control,'' in \emph{IEEE/RSJ International Conference on Intelligent Robots and Systems}.\hskip 1em plus 0.5em minus 0.4em\relax IEEE, 2012, pp. 5026--5033.

\bibitem{Silver2016MasteringTG}
D.~Silver, A.~Huang, C.~J. Maddison, A.~Guez, L.~Sifre, G.~van~den Driessche, J.~Schrittwieser, I.~Antonoglou, V.~Panneershelvam, M.~Lanctot, S.~Dieleman, D.~Grewe, J.~Nham, N.~Kalchbrenner, I.~Sutskever, T.~P. Lillicrap, M.~Leach, K.~Kavukcuoglu, T.~Graepel, and D.~Hassabis, ``Mastering the game of {Go} with deep neural networks and tree search,'' \emph{Nature}, vol. 529, pp. 484--489, 2016.

\bibitem{meta2022human}
FAIR, A.~Bakhtin, N.~Brown, E.~Dinan, G.~Farina, C.~Flaherty, D.~Fried, A.~Goff, J.~Gray, H.~Hu \emph{et~al.}, ``Human-level play in the game of diplomacy by combining language models with strategic reasoning,'' \emph{Science}, vol. 378, no. 6624, pp. 1067--1074, 2022.

\bibitem{wurman2022outracing}
P.~R. Wurman, S.~Barrett, K.~Kawamoto, J.~MacGlashan, K.~Subramanian, T.~J. Walsh, R.~Capobianco, A.~Devlic, F.~Eckert, F.~Fuchs, L.~Gilpin, P.~Khandelwal, V.~Kompella, H.~Lin, P.~MacAlpine, D.~Oller, T.~Seno, C.~Sherstan, M.~D. Thomure, H.~Aghabozorgi, L.~Barrett, R.~Douglas, D.~Whitehead, P.~Dürr, P.~Stone, M.~Spranger, and H.~Kitano, ``Outracing champion {Gran Turismo} drivers with deep reinforcement learning,'' \emph{Nature}, vol. 602, no. 7896, pp. 223--228, 2022.

\bibitem{segler2018planning}
M.~H. Segler, M.~Preuss, and M.~P. Waller, ``Planning chemical syntheses with deep neural networks and symbolic {AI},'' \emph{Nature}, vol. 555, no. 7698, pp. 604--610, 2018.

\bibitem{davies2021advancing}
A.~Davies, P.~Veli{\v{c}}kovi{\'c}, L.~Buesing, S.~Blackwell, D.~Zheng, N.~Toma{\v{s}}ev, R.~Tanburn, P.~Battaglia, C.~Blundell, A.~Juh{\'a}sz \emph{et~al.}, ``Advancing mathematics by guiding human intuition with {AI},'' \emph{Nature}, vol. 600, no. 7887, pp. 70--74, 2021.

\bibitem{duenez2023social}
E.~A. Du{\'e}{\~n}ez-Guzm{\'a}n, S.~Sadedin, J.~X. Wang, K.~R. McKee, and J.~Z. Leibo, ``A social path to human-like artificial intelligence,'' \emph{Nature Machine Intelligence}, vol.~5, no.~11, pp. 1181--1188, 2023.

\bibitem{fachada2021colorshapelinks}
N.~Fachada, ``{ColorShapeLinks}: A board game {AI} competition for educators and students,'' \emph{Computers and Education: Artificial Intelligence}, vol.~2, p. 100014, 2021.

\bibitem{mjai}
H.~Ichikawa, ``Mjai: {G}ame server for japanese mahjong {AI},'' \url{https://github.com/gimite/mjai}, 2014.

\bibitem{koyamada2022mjx}
S.~Koyamada, K.~Habara, N.~Goto, S.~Okano, S.~Nishimori, and S.~Ishii, ``Mjx: A framework for mahjong {AI} research,'' in \emph{IEEE Conference on Games}.\hskip 1em plus 0.5em minus 0.4em\relax IEEE, 2022, pp. 504--507.

\bibitem{justesen2019blood}
N.~Justesen, L.~M. Uth, C.~Jakobsen, P.~D. Moore, J.~Togelius, and S.~Risi, ``Blood bowl: A new board game challenge and competition for {AI},'' in \emph{IEEE Conference on Games}.\hskip 1em plus 0.5em minus 0.4em\relax IEEE, 2019, pp. 1--8.

\bibitem{bakhtin2023mastering}
\BIBentryALTinterwordspacing
A.~Bakhtin, D.~J. Wu, A.~Lerer, J.~Gray, A.~P. Jacob, G.~Farina, A.~H. Miller, and N.~Brown, ``Mastering the game of no-press diplomacy via human-regularized reinforcement learning and planning,'' in \emph{The Eleventh International Conference on Learning Representations}, 2023. [Online]. Available: \url{https://openreview.net/forum?id=F61FwJTZhb}
\BIBentrySTDinterwordspacing

\bibitem{reis2021vgc}
S.~Reis, L.~P. Reis, and N.~Lau, ``{VGC} {AI} competition-{A} new model of meta-game balance {AI} competition,'' in \emph{IEEE Conference on Games}.\hskip 1em plus 0.5em minus 0.4em\relax IEEE, 2021, pp. 1--8.

\bibitem{kowalski2023introducing}
J.~Kowalski, R.~Miernik, K.~Polak, D.~Budzki, and D.~Kowalik, ``Introducing {Tales of Tribute AI} competition,'' \emph{arXiv preprint arXiv:2305.08234}, 2023.

\bibitem{dockhorn2019introducing}
A.~Dockhorn and S.~Mostaghim, ``Introducing the {H}earthstone-{AI} competition,'' \emph{arXiv preprint arXiv:1906.04238}, pp. 1--4, 2019.

\bibitem{fireplace2014}
HearthSim, ``Fireplace,'' \url{https://github.com/jleclanche/fireplace/}, 2014.

\bibitem{Dali2022}
L.~Dali, ``Bridge engine,'' \url{https://github.com/lorserker/ben}, 2022.

\bibitem{bard2020hanabi}
N.~Bard, J.~N. Foerster, S.~Chandar, N.~Burch, M.~Lanctot, H.~F. Song, E.~Parisotto, V.~Dumoulin, S.~Moitra, E.~Hughes, I.~Dunning, S.~Mourad, H.~Larochelle, M.~G. Bellemare, and M.~Bowling, ``The {H}anabi challenge: {A} new frontier for {AI} research,'' \emph{Artificial Intelligence}, vol. 280, p. 103216, 2020.

\bibitem{barros2020s}
\BIBentryALTinterwordspacing
P.~Barros, A.~Sciutti, A.~C. Bloem, I.~M. Hootsmans, L.~M. Opheij, R.~H. Toebosch, and E.~Barakova, ``It's food fight! {D}esigning the chef's hat card game for affective-aware {HRI},'' in \emph{ACM/IEEE International Conference on Human-Robot Interaction}.\hskip 1em plus 0.5em minus 0.4em\relax Association for Computing Machinery, 2021, p. 524–528. [Online]. Available: \url{https://doi.org/10.1145/3434074.3447227}
\BIBentrySTDinterwordspacing

\bibitem{toriumi2016ai}
F.~Toriumi, H.~Osawa, M.~Inaba, D.~Katagami, K.~Shinoda, and H.~Matsubara, ``{AI} wolf contest—{D}evelopment of game {AI} using collective intelligence,'' in \emph{Computer Games}.\hskip 1em plus 0.5em minus 0.4em\relax Springer, 2016, pp. 101--115.

\bibitem{urbanek2019learning}
J.~Urbanek, A.~Fan, S.~Karamcheti, S.~Jain, S.~Humeau, E.~Dinan, T.~Rockt{\"a}schel, D.~Kiela, A.~Szlam, and J.~Weston, ``Learning to speak and act in a fantasy text adventure game,'' in \emph{Proceedings of the 2019 Conference on Empirical Methods in Natural Language Processing and the 9th International Joint Conference on Natural Language Processing}, 2019, p. 673–683.

\bibitem{brown2021snakes}
J.~A. Brown, L.~J.~P. de~Araujo, and A.~Grichshenko, ``Snakes {AI} competition 2020 and 2021 report,'' \emph{arXiv preprint arXiv:2108.05136}, 2021.

\bibitem{earle2020using}
S.~Earle, ``Using fractal neural networks to play {Simcity} 1 and {Conway's Game of Life} at variable scales,'' \emph{arXiv preprint arXiv:2002.03896}, 2020.

\bibitem{kuttler2020nethack}
H.~K{\"u}ttler, N.~Nardelli, A.~Miller, R.~Raileanu, M.~Selvatici, E.~Grefenstette, and T.~Rockt{\"a}schel, ``The {N}ethack learning environment,'' in \emph{International Conference on Neural Information Processing Systems}, vol.~33, 2020, pp. 7671--7684.

\bibitem{resnick2018pommerman}
C.~Resnick, W.~Eldridge, D.~Ha, D.~Britz, J.~Foerster, J.~Togelius, K.~Cho, and J.~Bruna, ``Pommerman: A multi-agent playground,'' \emph{CEUR Workshop}, vol. 2282, pp. 1--6, 2018.

\bibitem{suarez2021neural}
\BIBentryALTinterwordspacing
J.~Suarez, Y.~Du, C.~Zhu, I.~Mordatch, and P.~Isola, ``The neural {MMO} platform for massively multiagent research,'' in \emph{Thirty-fifth Conference on Neural Information Processing Systems Datasets and Benchmarks Track}, 2021. [Online]. Available: \url{https://openreview.net/forum?id=J0d-I8yFtP}
\BIBentrySTDinterwordspacing

\bibitem{ontanon2018first}
S.~Onta{\~n}{\'o}n, N.~A. Barriga, C.~R. Silva, R.~O. Moraes, and L.~H. Lelis, ``The first micro{RTS} artificial intelligence competition,'' \emph{AI Magazine}, vol.~39, no.~1, pp. 75--83, 2018.

\bibitem{perez2020tribes}
D.~Perez-Liebana, Y.-J. Hsu, S.~Emmanouilidis, B.~Khaleque, and R.~Gaina, ``Tribes: A new turn-based strategy game for {AI} research,'' in \emph{AAAI Conference on Artificial Intelligence and Interactive Digital Entertainment}, vol.~16, 2020, pp. 252--258.

\bibitem{carroll2019utility}
\BIBentryALTinterwordspacing
M.~Carroll, R.~Shah, M.~K. Ho, T.~L. Griffiths, S.~A. Seshia, P.~Abbeel, and A.~Dragan, ``On the utility of learning about humans for {Human-AI} coordination,'' in \emph{International Conference on Neural Information Processing Systems}, vol.~32.\hskip 1em plus 0.5em minus 0.4em\relax Curran Associates Inc., 2019. [Online]. Available: \url{https://proceedings.neurips.cc/paper_files/paper/2019/file/f5b1b89d98b7286673128a5fb112cb9a-Paper.pdf}
\BIBentrySTDinterwordspacing

\bibitem{gong2023mindagent}
R.~Gong, Q.~Huang, X.~Ma, H.~Vo, Z.~Durante, Y.~Noda, Z.~Zheng, S.-C. Zhu, D.~Terzopoulos, L.~Fei-Fei, and J.~Gao, ``{MindAgent}: Emergent gaming interaction,'' \emph{arXiv preprint arXiv: 2309.09971}, 2023.

\bibitem{karakovskiy2012mario}
S.~Karakovskiy and J.~Togelius, ``The {Mario AI} benchmark and competitions,'' \emph{IEEE Transactions on Computational Intelligence and {AI} in Games}, vol.~4, no.~1, pp. 55--67, 2012.

\bibitem{williams2016ms}
P.~R. Williams, D.~Perez-Liebana, and S.~M. Lucas, ``{Ms. Pac-Man Versus Ghost Team} {CIG} 2016 competition,'' in \emph{IEEE Conference on Computational Intelligence and Games}.\hskip 1em plus 0.5em minus 0.4em\relax IEEE, 2016, pp. 1--8.

\bibitem{prada2015geometry}
R.~Prada, P.~Lopes, J.~Catarino, J.~Quit{\'e}rio, and F.~S. Melo, ``The geometry friends game {AI} competition,'' in \emph{IEEE Conference on Computational Intelligence and Games}, 2015, pp. 431--438.

\bibitem{lucas2018game}
S.~M. Lucas, ``Game {AI} research with fast {Planet Wars} variants,'' in \emph{IEEE Conference on Computational Intelligence and Games}.\hskip 1em plus 0.5em minus 0.4em\relax IEEE, 2018, pp. 1--4.

\bibitem{arnett2024x}
T.~Arnett, K.~Cohen, N.~Ernest, Z.~Phillips, L.~Pickering, and S.~King, ``Explainable fuzzy {AI} challenge,'' \url{https://xfuzzycomp.github.io/XFC/index.html}, 2021.

\bibitem{khan2022darefightingice}
I.~Khan, T.~Van~Nguyen, X.~Dai, and R.~Thawonmas, ``{DareFightingICE} competition: A fighting game sound design and {AI} competition,'' in \emph{IEEE Conference on Games}.\hskip 1em plus 0.5em minus 0.4em\relax IEEE, 2022, pp. 478--485.

\bibitem{renz2019ai}
J.~Renz, X.~Ge, M.~Stephenson, and P.~Zhang, ``{AI} meets {Angry Birds},'' \emph{Nature Machine Intelligence}, vol.~1, no.~7, pp. 328--328, 2019.

\bibitem{nsh-github}
NeteaseFuXiRL, ``nsh: A simulated game environment for the commercial game justice online,'' \url{https://github.com/NeteaseFuxiRL/nsh}, 2020.

\bibitem{wei2022hok_env}
H.~Wei, J.~Chen, X.~Ji, H.~Qin, M.~Deng, S.~Li, L.~Wang, W.~Zhang, Y.~Yu, L.~Liu, L.~Huang, D.~Ye, Q.~Fu, and W.~Yang, ``{Honor of Kings Arena}: An environment for generalization in competitive reinforcement learning,'' in \emph{International Conference on Neural Information Processing Systems}, 2022, pp. 1--12.

\bibitem{FuXiRL2023DunkCityDynasty}
NeteaseFuXiRL, ``Dunk city dynasty,'' \url{https://github.com/FuxiRL/DunkCityDynasty}, 2023.

\bibitem{guss2019minerldata}
W.~H. Guss, B.~Houghton, N.~Topin, P.~Wang, C.~Codel, M.~Veloso, and R.~Salakhutdinov, ``Mine{RL}: A large-scale dataset of minecraft demonstrations,'' in \emph{International Joint Conference on Artificial Intelligence}, 2019, pp. 1--7.

\bibitem{fan2022minedojo}
L.~Fan, G.~Wang, Y.~Jiang, A.~Mandlekar, Y.~Yang, H.~Zhu, A.~Tang, D.-A. Huang, Y.~Zhu, and A.~Anandkumar, ``{MineDojo}: Building open-ended embodied agents with internet-scale knowledge,'' in \emph{Advances in Neural Information Processing Systems}, 2022, pp. 1--20.

\bibitem{kurach2020google}
K.~Kurach, A.~Raichuk, P.~Sta{\'n}czyk, M.~Zaj{\k{a}}c, O.~Bachem, L.~Espeholt, C.~Riquelme, D.~Vincent, M.~Michalski, O.~Bousquet, and S.~Gelly, ``Google research football: A novel reinforcement learning environment,'' in \emph{Proceedings of the AAAI Conference on Artificial Intelligence}, vol.~34, no.~4, 2020, pp. 4501--4510.

\bibitem{kempka2016vizdoom}
M.~Kempka, M.~Wydmuch, G.~Runc, J.~Toczek, and W.~Ja{\'s}kowski, ``{ViZDoom}: A doom-based {AI} research platform for visual reinforcement learning,'' in \emph{IEEE Conference on Computational Intelligence and Games}.\hskip 1em plus 0.5em minus 0.4em\relax IEEE, 2016, pp. 1--8.

\bibitem{baker2019emergent}
\BIBentryALTinterwordspacing
B.~Baker, I.~Kanitscheider, T.~Markov, Y.~Wu, G.~Powell, B.~McGrew, and I.~Mordatch, ``Emergent tool use from multi-agent autocurricula,'' in \emph{International Conference on Learning Representations}, 2020. [Online]. Available: \url{https://openreview.net/forum?id=SkxpxJBKwS}
\BIBentrySTDinterwordspacing

\bibitem{genesereth2005general}
M.~Genesereth, N.~Love, and B.~Pell, ``General game playing: Overview of the {AAAI} competition,'' \emph{AI Magazine}, vol.~26, no.~2, pp. 62--72, 2005.

\bibitem{zhou2017botzone}
H.~Zhou, Y.~Zhou, H.~Zhang, H.~Huang, and W.~Li, ``Botzone: A competitive and interactive platform for game {AI} education,'' in \emph{ACM Turing 50th Celebration Conference-China}, 2017, pp. 1--5.

\bibitem{lanctot2019openspiel}
M.~Lanctot, E.~Lockhart, J.-B. Lespiau, V.~Zambaldi, S.~Upadhyay, J.~P{\'e}rolat, S.~Srinivasan, F.~Timbers, K.~Tuyls, S.~Omidshafiei, D.~Hennes, D.~Morrill, P.~Muller, T.~Ewalds, R.~Faulkner, J.~Kramár, B.~De~Vylder, B.~Saeta, J.~Bradbury, D.~Ding, S.~Borgeaud, M.~Lai, J.~Schrittwieser, T.~Anthony, E.~Hughes, I.~Danihelka, and J.~Ryan-Davis, ``{OpenSpiel}: A framework for reinforcement learning in games,'' \emph{arXiv preprint arXiv:1908.09453}, 2019.

\bibitem{gaina2020tag}
R.~D. Gaina, M.~Balla, A.~Dockhorn, R.~Montoliu, and D.~Perez-Liebana, ``{TAG: A Tabletop Games Framework},'' in \emph{{Experimental AI in Games, AIIDE 2020 Workshop}}, 2020, pp. 1--7.

\bibitem{stephenson2019ludii}
M.~Stephenson, E.~Piette, D.~J. Soemers, and C.~Browne, ``Ludii as a competition platform,'' in \emph{IEEE Conference on Games}.\hskip 1em plus 0.5em minus 0.4em\relax IEEE, 2019, pp. 1--8.

\bibitem{koyamada2024pgx}
S.~Koyamada, S.~Okano, S.~Nishimori, Y.~Murata, K.~Habara, H.~Kita, and S.~Ishii, ``Pgx: Hardware-accelerated parallel game simulators for reinforcement learning,'' in \emph{Advances in Neural Information Processing Systems}, vol.~36, 2023, pp. 1--13.

\bibitem{brockman2016openai}
G.~Brockman, V.~Cheung, L.~Pettersson, J.~Schneider, J.~Schulman, J.~Tang, and W.~Zaremba, ``{OpenAI Gym},'' \emph{arXiv preprint arXiv:1606.01540}, 2016.

\bibitem{terry2021pettingzoo}
J.~K. Terry, B.~Black, A.~Hari, L.~S. Santos, C.~Dieffendahl, N.~L. Williams, Y.~Lokesh, C.~Horsch, and P.~Ravi, ``Pettingzoo: Gym for multi-agent reinforcement learning,'' in \emph{International Conference on Neural Information Processing Systems}, vol.~34, 2021, pp. 15\,032--15\,043.

\bibitem{perez2016general}
D.~Perez-Liebana, S.~Samothrakis, J.~Togelius, T.~Schaul, and S.~M. Lucas, ``General video game {AI}: Competition, challenges and opportunities,'' in \emph{Thirtieth AAAI Conference on Artificial Intelligence}, vol.~30, 2016, pp. 1--3.

\bibitem{perez2019general}
D.~Perez-Liebana, J.~Liu, A.~Khalifa, R.~D. Gaina, J.~Togelius, and S.~M. Lucas, ``General video game {AI}: A multitrack framework for evaluating agents, games, and content generation algorithms,'' \emph{IEEE Transactions on Games}, vol.~11, no.~3, pp. 195--214, 2019.

\bibitem{minigrid}
M.~Chevalier-Boisvert, L.~Willems, and S.~Pal, ``Minimalistic gridworld environment for gymnasium,'' \url{https://github.com/Farama-Foundation/Minigrid}, 2018.

\bibitem{leibo2021meltingpot}
J.~Z. Leibo, E.~A. Due{\~n}ez-Guzman, A.~Vezhnevets, J.~P. Agapiou, P.~Sunehag, R.~Koster, J.~Matyas, C.~Beattie, I.~Mordatch, and T.~Graepel, ``Scalable evaluation of multi-agent reinforcement learning with melting pot,'' in \emph{International Conference on Machine Learning}.\hskip 1em plus 0.5em minus 0.4em\relax PMLR, 2021, pp. 6187--6199.

\bibitem{dockhorn2020stratega}
A.~Dockhorn, J.~H. Grueso, D.~Jeurissen, and D.~P. Liebana, ``{STRATEGA: {A} General Strategy Games Framework},'' in \emph{{AAAI Conference on Artificial Intelligence and Interactive Digital Entertainment Workshop on Artificial Intelligence for Strategy Games}}, 2020, pp. 1--7.

\bibitem{bamford2021griddly}
C.~Bamford, ``Griddly: A platform for {AI} research in games,'' \emph{Software Impacts}, vol.~8, p. 100066, 2021.

\bibitem{bhonker2017playing}
\BIBentryALTinterwordspacing
N.~Bhonker, S.~Rozenberg, and I.~Hubara, ``Playing {SNES} in the retro learning environment,'' \emph{International Conference on Learning Representations Workshop}, 2017. [Online]. Available: \url{https://openreview.net/forum?id=HysBZSqlx}
\BIBentrySTDinterwordspacing

\bibitem{cobbe2020leveraging}
K.~Cobbe, C.~Hesse, J.~Hilton, and J.~Schulman, ``Leveraging procedural generation to benchmark reinforcement learning,'' in \emph{International Conference on Machine Learning}, 2020, pp. 2048--2056.

\bibitem{Zhang2022olympics}
H.~Zhang and Y.~Cui, ``{AI} {Olympics} competition,'' \url{https://github.com/jidiai/Competition_Olympics-Integrated}, 2022.

\bibitem{beattie2016deepmind}
C.~Beattie, J.~Z. Leibo, D.~Teplyashin, T.~Ward, M.~Wainwright, H.~K{\"{u}}ttler, A.~Lefrancq, S.~Green, V.~Vald{\'{e}}s, A.~Sadik, J.~Schrittwieser, K.~Anderson, S.~York, M.~Cant, A.~Cain, A.~Bolton, S.~Gaffney, H.~King, D.~Hassabis, S.~Legg, and S.~Petersen, ``{DeepMind lab},'' \emph{arXiv preprint arXiv:1612.03801}, 2016.

\bibitem{gym_miniworld}
M.~Chevalier-Boisvert, ``Miniworld: Minimalistic {3D} environment for {RL} $\&$ robotics research,'' \url{https://github.com/maximecb/gym-miniworld}, 2018.

\bibitem{juliani2018unity}
A.~Juliani, V.-P. Berges, E.~Teng, A.~Cohen, J.~Harper, C.~Elion, C.~Goy, Y.~Gao, H.~Henry, M.~Mattar, and D.~Lange, ``Unity: A general platform for intelligent agents,'' \emph{arXiv preprint arXiv:1809.02627}, 2018.

\bibitem{ventos2017bridge}
V.~Ventos, V.~Ventos, and O.~Teytaud, ``Le bridge, nouveau d{\'e}fi de l’intelligence artificielle?'' \emph{Revue des Sciences et Technologies de l'Information-S{\'e}rie RIA: Revue d'Intelligence Artificielle}, vol.~31, no.~3, pp. 249--279, 2017.

\bibitem{littman1994markov}
M.~L. Littman, ``Markov games as a framework for multi-agent reinforcement learning,'' in \emph{Machine Learning Proceedings 1994}.\hskip 1em plus 0.5em minus 0.4em\relax Elsevier, 1994, pp. 157--163.

\bibitem{bucsoniu2010multi}
L.~Bu{\c{s}}oniu, R.~Babu{\v{s}}ka, and B.~De~Schutter, ``Multi-agent reinforcement learning: An overview,'' \emph{Innovations in Multi-agent Systems and Applications-1}, pp. 183--221, 2010.

\bibitem{zhang2021multi}
K.~Zhang, Z.~Yang, and T.~Ba{\c{s}}ar, ``Multi-agent reinforcement learning: A selective overview of theories and algorithms,'' \emph{Handbook of Reinforcement Learning and Control}, pp. 321--384, 2021.

\bibitem{fudenberg1991game}
D.~Fudenberg and J.~Tirole, \emph{Game {T}heory}.\hskip 1em plus 0.5em minus 0.4em\relax MIT press, 1991.

\bibitem{hazra2022applications}
T.~Hazra and K.~Anjaria, ``Applications of game theory in deep learning: {A} survey,'' \emph{Multimedia Tools and Applications}, vol.~81, no.~6, pp. 8963--8994, 2022.

\bibitem{boyle2011role}
E.~Boyle, T.~M. Connolly, and T.~Hainey, ``The role of psychology in understanding the impact of computer games,'' \emph{Entertainment computing}, vol.~2, no.~2, pp. 69--74, 2011.

\bibitem{stephenson2019overview}
M.~Stephenson, E.~Piette, D.~J. Soemers, and C.~Browne, ``An overview of the {Ludii} general game system,'' in \emph{IEEE Conference on Games}.\hskip 1em plus 0.5em minus 0.4em\relax IEEE, 2019, pp. 1--2.

\bibitem{browne2010evolutionary}
C.~Browne and F.~Maire, ``Evolutionary game design,'' \emph{IEEE Transactions on Computational Intelligence and AI in Games}, vol.~2, no.~1, pp. 1--16, 2010.

\bibitem{mordatch2017emergence}
I.~Mordatch and P.~Abbeel, ``Emergence of grounded compositional language in multi-agent populations,'' in \emph{Proceedings of the AAAI Conference on Artificial Intelligence}.\hskip 1em plus 0.5em minus 0.4em\relax AAAI, 2018, pp. 1495–--1502.

\bibitem{schaul2013pyvgdl}
T.~Schaul, ``A video game description language for model-based or interactive learning,'' in \emph{IEEE Conference on Computational Intelligence in Games}.\hskip 1em plus 0.5em minus 0.4em\relax IEEE, 2013, pp. 1--8.

\bibitem{sutton2018reinforcement}
R.~S. Sutton and A.~G. Barto, \emph{Reinforcement {L}earning: An {I}ntroduction}.\hskip 1em plus 0.5em minus 0.4em\relax MIT press, 2018.

\bibitem{hussein2017imitation}
A.~Hussein, M.~M. Gaber, E.~Elyan, and C.~Jayne, ``Imitation learning: A survey of learning methods,'' \emph{ACM Computing Surveys (CSUR)}, vol.~50, no.~2, pp. 1--35, 2017.

\bibitem{browne2012survey}
C.~B. Browne, E.~Powley, D.~Whitehouse, S.~M. Lucas, P.~I. Cowling, P.~Rohlfshagen, S.~Tavener, D.~Perez, S.~Samothrakis, and S.~Colton, ``A survey of {Monte Carlo} tree search methods,'' \emph{IEEE Transactions on Computational Intelligence and AI in games}, vol.~4, no.~1, pp. 1--43, 2012.

\bibitem{swiechowski2023monte}
M.~{\'S}wiechowski, K.~Godlewski, B.~Sawicki, and J.~Ma{\'n}dziuk, ``{Monte Carlo} tree search: A review of recent modifications and applications,'' \emph{Artificial Intelligence Review}, vol.~56, no.~3, pp. 2497--2562, 2023.

\bibitem{bellemare2016unifying}
M.~Bellemare, S.~Srinivasan, G.~Ostrovski, T.~Schaul, D.~Saxton, and R.~Munos, ``Unifying count-based exploration and intrinsic motivation,'' in \emph{Advances in Neural Information Processing Systems}, vol.~29, 2016, pp. 1--9.

\bibitem{pathak2017curiosity}
D.~Pathak, P.~Agrawal, A.~A. Efros, and T.~Darrell, ``Curiosity-driven exploration by self-supervised prediction,'' in \emph{International Conference on Machine Learning}.\hskip 1em plus 0.5em minus 0.4em\relax PMLR, 2017, pp. 2778--2787.

\bibitem{osband2016deep}
I.~Osband, C.~Blundell, A.~Pritzel, and B.~Van~Roy, ``Deep exploration via bootstrapped {DQN},'' \emph{Advances in Neural Information Processing Systems}, vol.~29, pp. 1--9, 2016.

\bibitem{bertoli2006strong}
P.~Bertoli, A.~Cimatti, M.~Roveri, and P.~Traverso, ``Strong planning under partial observability,'' \emph{Artificial Intelligence}, vol. 170, no. 4-5, pp. 337--384, 2006.

\bibitem{kurniawati2022partially}
H.~Kurniawati, ``Partially observable markov decision processes and robotics,'' \emph{Annual Review of Control, Robotics, and Autonomous Systems}, vol.~5, pp. 253--277, 2022.

\bibitem{liebana2020general}
D.~P. Li{\'e}bana, S.~M. Lucas, R.~D. Gaina, J.~Togelius, A.~Khalifa, and J.~Liu, \emph{General video game artificial intelligence}.\hskip 1em plus 0.5em minus 0.4em\relax Springer, 2020.

\bibitem{lu2013fighting}
F.~Lu, K.~Yamamoto, L.~H. Nomura, S.~Mizuno, Y.~Lee, and R.~Thawonmas, ``Fighting game artificial intelligence competition platform,'' in \emph{Global Conference on Consumer Electronics}.\hskip 1em plus 0.5em minus 0.4em\relax IEEE, 2013, pp. 320--323.

\bibitem{jeon2023raidenv}
H.-C. Jeon, I.-C. Baek, C.-m. Bae, T.~Park, W.~You, T.~Ha, H.~Jung, J.~Noh, S.~Oh, and K.-J. Kim, ``{RaidEnv}: Exploring new challenges in automated content balancing for boss raid games,'' \emph{IEEE Transactions on Games}, pp. 1--14, 2023, (early access).

\bibitem{ferreira2014a}
L.~Ferreira and C.~Toledo, ``A search-based approach for generating {Angry Birds} levels,'' in \emph{Proceedings of the 9th IEEE International Conference on Computational Intelligence in Games}, 2014, pp. 1--8.

\bibitem{guzdial2019friend}
M.~Guzdial, N.~Liao, J.~Chen, S.-Y. Chen, S.~Shah, V.~Shah, J.~Reno, G.~Smith, and M.~O. Riedl, ``Friend, collaborator, student, manager: How design of an {AI}-driven game level editor affects creators,'' in \emph{CHI Conference on Human Factors in Computing Systems}, 2019, pp. 1--13.

\bibitem{hog2}
N.~Sturtevant, ``Hierarchical open graph 2,'' \url{https://github.com/nathansttt/hog2}, 2015.

\bibitem{heijne2017procedural}
N.~Heijne and S.~Bakkes, ``Procedural {Zelda}: A {PCG} environment for player experience research,'' in \emph{International Conference on the Foundations of Digital Games}, 2017, pp. 1--10.

\bibitem{khalifa2018talakat}
A.~Khalifa, S.~Lee, A.~Nealen, and J.~Togelius, ``Talakat: Bullet hell generation through constrained {MAP-Elites},'' in \emph{Genetic and Evolutionary Computation Conference}, 2018, pp. 1047--1054.

\bibitem{wang2021keiki}
Z.~Wang, J.~Liu, and G.~N. Yannakakis, ``Keiki: Towards realistic danmaku generation via sequential {GAN}s,'' in \emph{IEEE Conference on Games}.\hskip 1em plus 0.5em minus 0.4em\relax IEEE, 2021, pp. 01--04.

\bibitem{khalifa2020pcgrl}
A.~Khalifa, P.~Bontrager, S.~Earle, and J.~Togelius, ``{PCGRL}: Procedural content generation via reinforcement learning,'' in \emph{AAAI Conference on Artificial Intelligence and Interactive Digital Entertainment}, vol.~16, 2020, pp. 95--101.

\bibitem{earle2021learning}
S.~Earle, M.~Edwards, A.~Khalifa, P.~Bontrager, and J.~Togelius, ``Learning controllable content generators,'' in \emph{2021 IEEE Conference on Games}.\hskip 1em plus 0.5em minus 0.4em\relax IEEE, 2021, pp. 1--9.

\bibitem{salge2020ai}
C.~Salge, M.~C. Green, R.~Canaan, F.~Skwarski, R.~Fritsch, A.~Brightmoore, S.~Ye, C.~Cao, and J.~Togelius, ``The {AI} settlement generation challenge in {Minecraft},'' \emph{KI-K{\"u}nstliche Intelligenz}, vol.~34, no.~1, pp. 19--31, 2020.

\bibitem{charity2020say}
M.~Charity, D.~Rajesh, R.~Ombok, and L.~B. Soros, ``Say “sul sul!” to simsim, a sims-inspired platform for sandbox game {AI},'' in \emph{AAAI Conference on Artificial Intelligence and Interactive Digital Entertainment}, vol.~16, 2020, pp. 182--188.

\bibitem{cote2019textworld}
M.~C{\^{o}}t{\'{e}}, {\'{A}}.~K{\'{a}}d{\'{a}}r, X.~Yuan, B.~Kybartas, T.~Barnes, E.~Fine, J.~Moore, M.~J. Hausknecht, L.~E. Asri, M.~Adada, W.~Tay, and A.~Trischler, ``Textworld: A learning environment for text-based games,'' in \emph{Computer Games}.\hskip 1em plus 0.5em minus 0.4em\relax Springer, 2019, pp. 41--75.

\bibitem{ammanabrolu2020bringing}
P.~Ammanabrolu, W.~Cheung, D.~Tu, W.~Broniec, and M.~Riedl, ``Bringing stories alive: Generating interactive fiction worlds,'' in \emph{AAAI Conference on Artificial Intelligence and Interactive Digital Entertainment}, vol.~16, 2020, pp. 3--9.

\bibitem{StepMania}
``Stepmania,'' \url{https://github.com/stepmania/stepmania}, 2014.

\bibitem{osu}
Peppy, ``{OSU!}'' \url{https://osu.ppy.sh/home}, 2007.

\bibitem{ling2016latent}
W.~Ling, P.~Blunsom, E.~Grefenstette, K.~M. Hermann, T.~Ko{\v{c}}isk{\`y}, F.~Wang, and A.~Senior, ``Latent predictor networks for code generation,'' in \emph{Proceedings of the 54th Annual Meeting of the Association for Computational Linguistics}, vol.~1, 2016, pp. 599--609.

\bibitem{santos2017monte}
A.~Santos, P.~A. Santos, and F.~S. Melo, ``{Monte Carlo} tree search experiments in {H}earthstone,'' in \emph{IEEE Conference on Computational Intelligence and Games}.\hskip 1em plus 0.5em minus 0.4em\relax IEEE, 2017, pp. 272--279.

\bibitem{guana2014phydsl}
V.~Guana and E.~Stroulia, ``Phy{DSL}: {A} code-generation environment for {2D} physics-based games,'' in \emph{IEEE Games, Entertainment, and Media Conference}, 2014, pp. 1--13.

\bibitem{holmgaard2014evolving}
C.~Holmg{\aa}rd, A.~Liapis, J.~Togelius, and G.~N. Yannakakis, ``Evolving personas for player decision modeling,'' in \emph{2014 IEEE Conference on Computational Intelligence and Games}.\hskip 1em plus 0.5em minus 0.4em\relax IEEE, 2014, pp. 1--8.

\bibitem{James2016vglc}
A.~J. Summerville, S.~Snodgrass, M.~Mateas, and S.~Ontan{\'o}n, ``The {VGLC}: The video game level corpus,'' \emph{7th Workshop on Procedural Content Generation}, pp. 1--6, 2016.

\bibitem{charity2020baba}
M.~Charity, A.~Khalifa, and J.~Togelius, ``Baba is y'all: Collaborative mixed-initiative level design,'' in \emph{IEEE Conference on Games}.\hskip 1em plus 0.5em minus 0.4em\relax IEEE, 2020, pp. 542--549.

\bibitem{bhaumik2021lode}
D.~Bhaumik, A.~Khalifa, and J.~Togelius, ``{Lode Encoder}: {AI}-constrained co-creativity,'' in \emph{IEEE Conference on Games}.\hskip 1em plus 0.5em minus 0.4em\relax IEEE, 2021, pp. 1--8.

\bibitem{ishihara2018monte}
M.~Ishihara, S.~Ito, R.~Ishii, T.~Harada, and R.~Thawonmas, ``Monte-carlo tree search for implementation of dynamic difficulty adjustment fighting game {AI}s having believable behaviors,'' in \emph{IEEE Conference on Computational Intelligence and Games}.\hskip 1em plus 0.5em minus 0.4em\relax IEEE, 2018, pp. 1--8.

\bibitem{liu2017evolving}
J.~Liu, J.~Togelius, D.~P{\'e}rez-Li{\'e}bana, and S.~M. Lucas, ``Evolving game skill-depth using general video game {AI} agents,'' in \emph{IEEE Congress on Evolutionary Computation}.\hskip 1em plus 0.5em minus 0.4em\relax IEEE, 2017, pp. 2299--2307.

\bibitem{wang2024negatively}
\BIBentryALTinterwordspacing
Z.~Wang, C.~Hu, J.~Liu, and X.~Yao, ``Negatively correlated ensemble reinforcement learning for online diverse game level generation,'' in \emph{The Twelfth International Conference on Learning Representations}, 2024. [Online]. Available: \url{https://openreview.net/forum?id=iAW2EQXfwb}
\BIBentrySTDinterwordspacing

\bibitem{sturtevant2020unexpected}
N.~Sturtevant, N.~Decroocq, A.~Tripodi, and M.~Guzdial, ``The unexpected consequence of incremental design changes,'' in \emph{Proceedings of the AAAI Conference on Artificial Intelligence and Interactive Digital Entertainment}, vol.~16, no.~1, 2020, pp. 130--136.

\bibitem{sturtevant2018exhaustive}
N.~Sturtevant and M.~Ota, ``Exhaustive and semi-exhaustive procedural content generation,'' in \emph{Proceedings of the AAAI Conference on Artificial Intelligence and Interactive Digital Entertainment}, vol.~14, no.~1, 2018, pp. 109--115.

\bibitem{chang2023survey}
Y.~Chang, X.~Wang, J.~Wang, Y.~Wu, L.~Yang, K.~Zhu, H.~Chen, X.~Yi, C.~Wang, Y.~Wang \emph{et~al.}, ``A survey on evaluation of large language models,'' \emph{ACM Transactions on Intelligent Systems and Technology}, vol.~15, no.~3, pp. 1--45, 2024.

\bibitem{gallotta2024large}
R.~Gallotta, G.~Todd, M.~Zammit, S.~Earle, A.~Liapis, J.~Togelius, and G.~N. Yannakakis, ``Large language models and games: A survey and roadmap,'' \emph{arXiv preprint arXiv:2402.18659}, 2024.

\bibitem{Abdullah2024ChatGPT4PCG}
F.~Abdullah, P.~Taveekitworachai, M.~F. Dewantoro, R.~Thawonmas, J.~Togelius, and J.~Renz, ``The 1st {ChatGPT4PCG} competition,'' \emph{IEEE Transactions on Games}, pp. 1--17, 2024, (early access).

\bibitem{salge2018generative}
C.~Salge, M.~C. Green, R.~Canaan, and J.~Togelius, ``Generative design in {Minecraft} ({GDMC}) settlement generation competition,'' in \emph{Proceedings of the 13th International Conference on the Foundations of Digital Games}, 2018, pp. 1--10.

\bibitem{huang2023generating}
S.~Huang, C.~Hu, J.~Togelius, and J.~Liu, ``Generating redstone style cities in {Minecraft},'' in \emph{IEEE Conference on Games}.\hskip 1em plus 0.5em minus 0.4em\relax IEEE, 2023, pp. 1--4.

\bibitem{chen2022analysis}
C.~Chen, Y.~Dai, J.~Poon, and C.~Han, ``An analysis of deep reinforcement learning agents for text-based games,'' \emph{arXiv preprint arXiv:2209.04105}, 2022.

\bibitem{donahue2017dance}
C.~Donahue, Z.~C. Lipton, and J.~McAuley, ``Dance dance convolution,'' in \emph{International Conference on Machine Learning}.\hskip 1em plus 0.5em minus 0.4em\relax PMLR, 2017, pp. 1039--1048.

\bibitem{franks2023ordinal}
\BIBentryALTinterwordspacing
B.~J. Franks, B.~Dinkelmann, M.~Kloft, and S.~Fellenz, ``Ordinal regression for difficulty prediction of {StepMania} levels,'' in \emph{Machine Learning and Knowledge Discovery in Databases: Applied Data Science and Demo Track: European Conference, ECML PKDD 2023, Turin, Italy, September 18–22, 2023, Proceedings, Part VI}.\hskip 1em plus 0.5em minus 0.4em\relax Berlin, Heidelberg: Springer-Verlag, 2023, p. 497–512. [Online]. Available: \url{https://doi.org/10.1007/978-3-031-43427-3_30}
\BIBentrySTDinterwordspacing

\bibitem{liang2019procedural}
Y.~Liang, W.~Li, and K.~Ikeda, ``Procedural content generation of rhythm games using deep learning methods,'' in \emph{Entertainment Computing and Serious Games}.\hskip 1em plus 0.5em minus 0.4em\relax Springer, 2019, pp. 134--145.

\bibitem{halina2021taikonation}
E.~Halina and M.~Guzdial, ``{TaikoNation}: Patterning-focused chart generation for rhythm action games,'' in \emph{Proceedings of the 16th International Conference on the Foundations of Digital Games}, 2021, pp. 1--10.

\bibitem{garcia2016evolutionary}
P.~Garc{\'\i}a-S{\'a}nchez, A.~Tonda, G.~Squillero, A.~Mora, and J.~J. Merelo, ``Evolutionary deckbuilding in {H}earthstone,'' in \emph{IEEE Conference on Computational Intelligence and Games}.\hskip 1em plus 0.5em minus 0.4em\relax IEEE, 2016, pp. 1--8.

\bibitem{zilio2018neural}
F.~Zilio, M.~Prates, and L.~Lamb, ``Neural networks models for analyzing {Magic: the Gathering} cards,'' in \emph{Proceedings of the 25th International Conference on Neural Information Processing}.\hskip 1em plus 0.5em minus 0.4em\relax Springer, 2018, pp. 227--239.

\bibitem{churchill2019magic}
A.~Churchill, S.~Biderman, and A.~Herrick, ``{Magic: The Gathering} is turing complete,'' in \emph{International Conference on Fun with Algorithms}, vol. 157.\hskip 1em plus 0.5em minus 0.4em\relax Schloss Dagstuhl--Leibniz-Zentrum f{\"u}r Informatik, 2020, pp. 9:1--9:19.

\bibitem{gansner2000open}
E.~R. Gansner and S.~C. North, ``An open graph visualization system and its applications to software engineering,'' \emph{Software: Practice and Experience}, vol.~30, no.~11, pp. 1203--1233, 2000.

\bibitem{reis2023adversarial}
S.~Reis, R.~Novais, L.~P. Reis, and N.~Lau, ``An adversarial approach for automated {Pokémon} team building and meta-game balance,'' \emph{IEEE Transactions on Games}, pp. 1--11, 2023, (early access).

\bibitem{khan2023fighting}
I.~Khan, T.~Van~Nguyen, C.~Nimpattanavong, and R.~Thawonmas, ``Fighting game adaptive background music for improved gameplay,'' in \emph{2023 IEEE Conference on Games}.\hskip 1em plus 0.5em minus 0.4em\relax IEEE, 2023, pp. 1--2.

\bibitem{khan2024enhanced}
I.~Khan, C.~Nimpattanavong, T.~Van~Nguyen, K.~Plupattanakit, and R.~Thawonmas, ``Enhanced {DareFightingICE} competitions: Sound design and {AI} competitions,'' \emph{arXiv preprint arXiv:2403.02687}, 2024.

\bibitem{shaker2012evolving}
N.~Shaker, M.~Nicolau, G.~N. Yannakakis, J.~Togelius, and M.~O'neill, ``Evolving levels for {Super Mario Bros} using grammatical evolution,'' in \emph{IEEE Conference on Computational Intelligence and Games}.\hskip 1em plus 0.5em minus 0.4em\relax IEEE, 2012, pp. 304--311.

\bibitem{gravina2018quality}
D.~Gravina, A.~Liapis, and G.~N. Yannakakis, ``Quality diversity through surprise,'' \emph{IEEE Transactions on Evolutionary Computation}, vol.~23, no.~4, pp. 603--616, 2018.

\bibitem{shu2021experience}
T.~Shu, J.~Liu, and G.~N. Yannakakis, ``Experience-driven {PCG} via reinforcement learning: {A} {Super Mario Bros} study,'' in \emph{IEEE Conference on Games}.\hskip 1em plus 0.5em minus 0.4em\relax IEEE, 2021, pp. 1--9.

\bibitem{volz2018evolving}
V.~Volz, J.~Schrum, J.~Liu, S.~M. Lucas, A.~Smith, and S.~Risi, ``Evolving mario levels in the latent space of a deep convolutional generative adversarial network,'' in \emph{Genetic and Evolutionary Computation Conference}, 2018, pp. 221--228.

\bibitem{dai2024procedural}
S.~Dai, X.~Zhu, N.~Li, T.~Dai, and Z.~Wang, ``Procedural level generation with diffusion models from a single example,'' in \emph{Proceedings of the AAAI Conference on Artificial Intelligence}, vol.~38, no.~9, 2024, pp. 10\,021--10\,029.

\bibitem{gonzalez2020finding}
M.~Gonz{\'a}lez-Duque, R.~B. Palm, D.~Ha, and S.~Risi, ``Finding game levels with the right difficulty in a few trials through intelligent trial-and-error,'' in \emph{2020 IEEE Conference on Games}.\hskip 1em plus 0.5em minus 0.4em\relax IEEE, 2020, pp. 503--510.

\bibitem{fontaine2019mapping}
M.~C. Fontaine, S.~Lee, L.~B. Soros, F.~de~Mesentier~Silva, J.~Togelius, and A.~K. Hoover, ``Mapping hearthstone deck spaces through map-elites with sliding boundaries,'' in \emph{Proceedings of The Genetic and Evolutionary Computation Conference}, 2019, pp. 161--169.

\bibitem{kunanusont2017n}
K.~Kunanusont, R.~D. Gaina, J.~Liu, D.~Perez-Liebana, and S.~M. Lucas, ``The n-tuple bandit evolutionary algorithm for automatic game improvement,'' in \emph{2017 IEEE Congress on Evolutionary Computation}.\hskip 1em plus 0.5em minus 0.4em\relax IEEE, 2017, pp. 2201--2208.

\bibitem{watkins1992q}
C.~J. Watkins and P.~Dayan, ``Q-learning,'' \emph{Machine Learning}, vol.~8, pp. 279--292, 1992.

\bibitem{dulac2015deep}
G.~Dulac-Arnold, R.~Evans, H.~van Hasselt, P.~Sunehag, T.~Lillicrap, J.~Hunt, T.~Mann, T.~Weber, T.~Degris, and B.~Coppin, ``Deep reinforcement learning in large discrete action spaces,'' \emph{arXiv preprint arXiv:1512.07679}, 2015.

\bibitem{burda2018exploration}
\BIBentryALTinterwordspacing
Y.~Burda, H.~Edwards, A.~Storkey, and O.~Klimov, ``Exploration by random network distillation,'' in \emph{International Conference on Learning Representations}, 2019. [Online]. Available: \url{https://openreview.net/forum?id=H1lJJnR5Ym}
\BIBentrySTDinterwordspacing

\bibitem{burda2018largescale}
\BIBentryALTinterwordspacing
Y.~Burda, H.~Edwards, D.~Pathak, A.~Storkey, T.~Darrell, and A.~A. Efros, ``Large-scale study of curiosity-driven learning,'' in \emph{International Conference on Learning Representations}, 2019. [Online]. Available: \url{https://openreview.net/forum?id=rJNwDjAqYX}
\BIBentrySTDinterwordspacing

\bibitem{conti2018improving}
E.~Conti, V.~Madhavan, F.~Petroski~Such, J.~Lehman, K.~Stanley, and J.~Clune, ``Improving exploration in evolution strategies for deep reinforcement learning via a population of novelty-seeking agents,'' in \emph{Advances in Neural Information Processing Systems}, vol.~31, 2018, pp. 1--12.

\bibitem{majid2023deep}
A.~Y. Majid, S.~Saaybi, V.~Francois-Lavet, R.~V. Prasad, and C.~Verhoeven, ``Deep reinforcement learning versus evolution strategies: {A} comparative survey,'' \emph{IEEE Transactions on Neural Networks and Learning Systems}, pp. 1--19, 2023, (early access).

\bibitem{fortunato2018noisy}
\BIBentryALTinterwordspacing
M.~Fortunato, M.~G. Azar, B.~Piot, J.~Menick, M.~Hessel, I.~Osband, A.~Graves, V.~Mnih, R.~Munos, D.~Hassabis, O.~Pietquin, C.~Blundell, and S.~Legg, ``Noisy networks for exploration,'' in \emph{International Conference on Learning Representations}, 2018. [Online]. Available: \url{https://openreview.net/forum?id=rywHCPkAW}
\BIBentrySTDinterwordspacing

\bibitem{jaakkola1994reinforcement}
T.~Jaakkola, S.~Singh, and M.~Jordan, ``Reinforcement learning algorithm for partially observable markov decision problems,'' \emph{Advances in Neural Information Processing Systems}, vol.~7, pp. 1--8, 1994.

\bibitem{kirk2023survey}
R.~Kirk, A.~Zhang, E.~Grefenstette, and T.~Rockt{\"a}schel, ``A survey of zero-shot generalisation in deep reinforcement learning,'' \emph{Journal of Artificial Intelligence Research}, vol.~76, pp. 201--264, 2023.

\bibitem{zhao2023survey}
W.~X. Zhao, K.~Zhou, J.~Li, T.~Tang, X.~Wang, Y.~Hou, Y.~Min, B.~Zhang, J.~Zhang, Z.~Dong \emph{et~al.}, ``A survey of large language models,'' \emph{arXiv preprint arXiv:2303.18223}, 2023.

\bibitem{openai2023chatgpt}
OpenAI, ``Introducing {ChatGPT},'' \url{https://openai.com/blog/chatgpt}, 2022.

\bibitem{xu2023exploring}
Y.~Xu, S.~Wang, P.~Li, F.~Luo, X.~Wang, W.~Liu, and Y.~Liu, ``Exploring large language models for communication games: An empirical study on {Werewolf},'' \emph{arXiv preprint arXiv:2309.04658}, 2023.

\bibitem{politowski2021survey}
C.~Politowski, F.~Petrillo, and Y.-G. Gu{\'e}h{\'e}neuc, ``A survey of video game testing,'' in \emph{IEEE/ACM International Conference on Automation of Software Test}.\hskip 1em plus 0.5em minus 0.4em\relax IEEE, 2021, pp. 90--99.

\bibitem{fraser2011evosuite}
G.~Fraser and A.~Arcuri, ``Evosuite: {A}utomatic test suite generation for object-oriented software,'' in \emph{Proceedings of the 19th ACM SIGSOFT Symposium and the 13th European Conference on Foundations of Software Engineering}, 2011, pp. 416--419.

\bibitem{zheng2019wuji}
Y.~Zheng, X.~Xie, T.~Su, L.~Ma, J.~Hao, Z.~Meng, Y.~Liu, R.~Shen, Y.~Chen, and C.~Fan, ``Wuji: {A}utomatic online combat game testing using evolutionary deep reinforcement learning,'' in \emph{34th IEEE/ACM International Conference on Automated Software Engineering}.\hskip 1em plus 0.5em minus 0.4em\relax IEEE, 2019, pp. 772--784.

\bibitem{bergdahl2020augmenting}
J.~Bergdahl, C.~Gordillo, K.~Tollmar, and L.~Gissl{\'e}n, ``Augmenting automated game testing with deep reinforcement learning,'' in \emph{IEEE Conference on Games}.\hskip 1em plus 0.5em minus 0.4em\relax IEEE, 2020, pp. 600--603.

\bibitem{pan2009survey}
S.~J. Pan and Q.~Yang, ``A survey on transfer learning,'' \emph{IEEE Transactions on Knowledge and Data Engineering}, vol.~22, no.~10, pp. 1345--1359, 2009.

\bibitem{preuss2018integrated}
M.~Preuss, T.~Pfeiffer, V.~Volz, and N.~Pflanzl, ``Integrated balancing of an {RTS} game: Case study and toolbox refinement,'' in \emph{IEEE Conference on Computational Intelligence and Games}.\hskip 1em plus 0.5em minus 0.4em\relax IEEE, 2018, pp. 1--8.

\bibitem{ferreira2020computer}
L.~Ferreira, L.~Lelis, and J.~Whitehead, ``Computer-generated music for tabletop role-playing games,'' in \emph{Proceedings of the AAAI Conference on Artificial Intelligence and Interactive Digital Entertainment}, vol.~16, no.~1, 2020, pp. 59--65.

\bibitem{hong2019game}
S.~Hong, S.~Kim, and S.~Kang, ``Game sprite generator using a multi discriminator {GAN},'' \emph{KSII Transactions on Internet \& Information Systems}, vol.~13, no.~8, pp. 4255--4269, 2019.

\bibitem{liu2024comprehensive}
J.~Liu, X.~Huang, T.~Huang, L.~Chen, Y.~Hou, S.~Tang, Z.~Liu, W.~Ouyang, W.~Zuo, J.~Jiang, and X.~Liu, ``A comprehensive survey on {3D} content generation,'' \emph{arXiv preprint arXiv:2402.01166}, 2024.

\bibitem{barth2023ghostwriter}
R.~Barth, ``The convergence of {AI} and creativity: {I}ntroducing ghostwriter,'' \url{https://news.ubisoft.com/en-us/article/7Cm07zbBGy4Xml6WgYi25d/the-convergence-of-ai-and-creativity-introducing-ghostwriter}, 2023.

\bibitem{kim2020learning}
S.~W. Kim, Y.~Zhou, J.~Philion, A.~Torralba, and S.~Fidler, ``Learning to simulate dynamic environments with {GameGAN},'' in \emph{IEEE/CVF Conference on Computer Vision and Pattern Recognition}, 2020, pp. 1231--1240.

\bibitem{shaker2013ropossum}
N.~Shaker, M.~Shaker, and J.~Togelius, ``Ropossum: An authoring tool for designing, optimizing and solving {Cut the Rope} levels,'' in \emph{Ninth Artificial Intelligence and Interactive Digital Entertainment Conference}, vol.~9, no.~1.\hskip 1em plus 0.5em minus 0.4em\relax AAAI, 2013, pp. 215--216.

\bibitem{beeching2021godot}
E.~Beeching, J.~Dibangoye, O.~Simonin, and C.~Wolf, ``Godot reinforcement learning agents,'' \emph{arXiv preprint arXiv:2112.03636.}, 2021.

\bibitem{johansen2019video}
M.~Johansen, M.~Pichlmair, and S.~Risi, ``Video game description language environment for unity machine learning agents,'' in \emph{IEEE Conference on Games}.\hskip 1em plus 0.5em minus 0.4em\relax IEEE, 2019, pp. 1--8.

\bibitem{qian2023communicative}
C.~Qian, X.~Cong, W.~Liu, C.~Yang, W.~Chen, Y.~Su, Y.~Dang, J.~Li, J.~Xu, D.~Li, Z.~Liu, and M.~Sun, ``Communicative agents for software development,'' \emph{arXiv preprint arXiv:2307.07924}, 2023.

\bibitem{puig2021watchandhelp}
\BIBentryALTinterwordspacing
X.~Puig, T.~Shu, S.~Li, Z.~Wang, Y.-H. Liao, J.~B. Tenenbaum, S.~Fidler, and A.~Torralba, ``Watch-and-help: A challenge for social perception and human-{AI} collaboration,'' in \emph{International Conference on Learning Representations}, 2021. [Online]. Available: \url{https://openreview.net/forum?id=w_7JMpGZRh0}
\BIBentrySTDinterwordspacing

\bibitem{Bartolomé2011can}
N.~A. Bartolomé, A.~M. Zorrilla, and B.~G. Zapirain, ``Can game-based therapies be trusted? is game-based education effective? {A} systematic review of the serious games for health and education,'' in \emph{16th International Conference on Computer Games}, 2011, pp. 275--282.

\bibitem{li2013game}
M.-C. Li and C.-C. Tsai, ``Game-based learning in science education: A review of relevant research,'' \emph{Journal of Science Education and Technology}, vol.~22, pp. 877--898, 2013.

\bibitem{abiyev2016brain}
R.~H. Abiyev, N.~Akkaya, E.~Aytac, I.~G{\"u}nsel, A.~{\c{C}}a{\u{g}}man \emph{et~al.}, ``Brain-computer interface for control of wheelchair using fuzzy neural networks,'' \emph{BioMed Research International}, vol. 2016, pp. 1--9, 2016.

\bibitem{brezinka2014computer}
V.~Brezinka, ``Computer games supporting cognitive behaviour therapy in children,'' \emph{Clinical Child Psychology and Psychiatry}, vol.~19, no.~1, pp. 100--110, 2014.

\bibitem{liu2020using}
Z.-Y. Liu, Z.~A. Shaikh, and F.~Gazizova, ``Using the concept of game-based learning in education.'' \emph{International Journal of Emerging Technologies in Learning}, vol.~15, pp. 1--12, 2020.

\bibitem{zhao2024playing}
Y.~Zhao, C.~Hu, and J.~Liu, ``Playing with {Monte-Carlo} tree search,'' \emph{IEEE Computational Intelligence Magazine}, vol.~19, no.~1, pp. 85--86, 2024.

\bibitem{togelius2024choose}
J.~Togelius and G.~N. Yannakakis, ``Choose your weapon: Survival strategies for depressed {AI} academics [point of view],'' \emph{Proceedings of the IEEE}, vol. 112, no.~1, pp. 4--11, 2024.

\end{thebibliography}


\end{document}